%% file: arxiv.tex
\documentclass{article} 

\newif\ificlrstyle \iclrstylefalse
\newif\ifanon      \anonfalse

\input{math_commands.tex}

\input{preamble/common}
\input{preamble/style-edison}

\input{preamble/anon}

\begin{document}

\input{frontmatter}

\normalsize

\input{sections/01-main/sec-01-introduction}

\input{sections/01-main/sec-02-related-work}

\input{sections/01-main/sec-03-method}

\input{sections/01-main/sec-04-experiments}

\input{sections/01-main/sec-05-limitations-and-future-work}

\input{sections/01-main/sec-06-conclusion}

\input{sections/01-main/sec-07-statement}

\bibliography{arxiv}
\bibliographystyle{iclr2027_conference}

\newpage
\appendix
\section{Appendix}

\input{sections/02-appendix/01-tasks-and-evaluation-details}

\input{sections/02-appendix/02-training-hyperparameters}

\input{sections/02-appendix/03-relative-errors}

\input{sections/02-appendix/04-haiku-judge-reward-evaluation-analysis}

\input{sections/02-appendix/05-qualitative-nih-grant-abstract-examples}

\end{document}

%% file: math_commands.tex
\usepackage{amsmath,amsfonts,bm}

\def\eqref#1{equation~\ref{#1}}

\def\1{\bm{1}}

\DeclareMathAlphabet{\mathsfit}{\encodingdefault}{\sfdefault}{m}{sl}
\SetMathAlphabet{\mathsfit}{bold}{\encodingdefault}{\sfdefault}{bx}{n}



%% file: preamble/common.tex
\usepackage{amsmath}
\usepackage{amssymb}
\usepackage{dsfont}

\usepackage{graphicx}
\usepackage{float}      
\usepackage{afterpage}

\usepackage{algorithm}
\usepackage{algpseudocode}
\usepackage{natbib}

\usepackage{tabularx}   
\usepackage{booktabs}   

\usepackage{xcolor}
\usepackage{tcolorbox}
\tcbuselibrary{skins,listings,breakable}
\usepackage{fvextra}

\usepackage{hyperref}
\hypersetup{
  colorlinks=true,
  linkcolor=black,
  citecolor=black,
  urlcolor=blue
}

\DeclareUnicodeCharacter{03BC}{\ensuremath{\mu}}
\DeclareUnicodeCharacter{03C7}{\ensuremath{\chi}}
\DeclareUnicodeCharacter{207B}{\ensuremath{^{-}}}
\DeclareUnicodeCharacter{2076}{\ensuremath{^{6}}}
\DeclareUnicodeCharacter{2082}{\ensuremath{_{2}}}
\DeclareUnicodeCharacter{2083}{\ensuremath{_{3}}}
\DeclareUnicodeCharacter{2086}{\ensuremath{_{6}}}
\DeclareUnicodeCharacter{2088}{\ensuremath{_{8}}}

\DefineVerbatimEnvironment{PromptBlock}{Verbatim}{
  fontsize=\small,
  vspace=0pt,
  breaklines=true,
  breakanywhere=false,
  breakautoindent=false,
  breakindent=0pt,
  breaksymbolleft={},
  breaksymbolright={},
  frame=single,
  framerule=0.3pt,
  framesep=6pt,
  rulecolor=\color{gray!35},
  bgcolor=gray!5
}

\definecolor{edisongray}{HTML}{f0f0f0}

\newenvironment{headerbox}{
  \begin{tcolorbox}[
    colback=edisongray,
    colframe=edisongray,
    frame hidden,
    arc=4mm,
    boxrule=0pt,
    boxsep=0pt,
    left=15pt,
    right=15pt,
    top=10pt,
    bottom=15pt,
    nobeforeafter
  ]
}{
  \end{tcolorbox}
}

\newtcolorbox{LLMOutput}[1]{
  enhanced,
  breakable,
  colback=gray!5,
  colframe=gray!35,
  colbacktitle=gray!12,
  coltitle=black,
  title={#1},
  fonttitle=\small\bfseries,
  fontupper=\small,
  boxrule=0.3pt,
  arc=0pt,
  boxsep=0pt,
  left=8pt,
  right=8pt,
  top=6pt,
  bottom=6pt,
  before skip=10pt,
  after skip=10pt,
  before upper={
    \setlength{\parindent}{0pt}
    \setlength{\parskip}{6pt}
  }
}

\definecolor{linkblue}{HTML}{1565C0}

\usepackage{placeins}

%% file: preamble/style-edison.tex
\usepackage{lmodern}

\usepackage{sectsty}

\definecolor{edisongray}{HTML}{f0f0f0}
\usepackage[left=0.9in, right=0.9in, top=0.85in, bottom=0.9in]{geometry}

\usepackage{hyperref}
\hypersetup{
  colorlinks=true,
  linkcolor=linkblue,
  citecolor=linkblue,
  urlcolor=linkblue
}
\input{preamble/cleveref}
\sectionfont{\sffamily}


%% file: preamble/cleveref.tex
\usepackage[nameinlink]{cleveref}

\crefname{section}{\S}{\S\S}
\Crefname{section}{Section}{Sections}
\crefname{subsection}{\S}{\S\S}
\Crefname{subsection}{Section}{Sections}
\crefname{subsubsection}{\S}{\S\S}
\Crefname{subsubsection}{Section}{Sections}
\crefformat{section}{\S#2#1#3}
\crefformat{subsection}{\S#2#1#3}
\crefformat{subsubsection}{\S#2#1#3}
\crefrangeformat{section}{\S\S#3#1#4--#5#2#6}
\crefrangeformat{subsection}{\S\S#3#1#4--#5#2#6}
\crefmultiformat{section}{\S\S#2#1#3}{ and #2#1#3}{, #2#1#3}{, and #2#1#3}
\crefmultiformat{subsection}{\S\S#2#1#3}{ and #2#1#3}{, #2#1#3}{, and #2#1#3}

\crefname{appendix}{App.}{Apps.}
\Crefname{appendix}{App.}{Apps.}
\crefname{subappendix}{App.}{Apps.}
\Crefname{subappendix}{App.}{Apps.}
\crefname{subsubappendix}{App.}{Apps.}
\Crefname{subsubappendix}{App.}{Apps.}

\crefname{figure}{Fig.}{Figs.}
\Crefname{figure}{Fig.}{Figs.}
\crefname{table}{Table}{Tables}
\Crefname{table}{Table}{Tables}
\crefname{equation}{Eq.}{Eqs.}
\Crefname{equation}{Eq.}{Eqs.}

%% file: preamble/anon.tex
\ifanon
\else
  
  \newcommand{\repohandle}[1]{\texttt{#1}}
\fi

\newcommand{\anontext}[2]{\ifanon\else#2\fi}

%% file: frontmatter.tex

\newcommand{\papertitle}{Learning to Reason from Reference Samples Alone}
\newcommand{\papersubtitle}{Reinforced Adversarial Learning for Modern LLM Post-Training}

\newcommand{\abstracttext}{%
In many settings where LLM post-training is desirable, high-quality answers are recognizable but difficult to score with explicit rewards. Reinforcement learning with verifiable rewards (RLVR) is limited to tasks where an explicit correctness signal is available. Prior adversarial approaches to training reasoning LLMs either complement verifiable rewards with co-trained critics or require reference reasoning traces. We ask whether a co-trained LLM judge can turn reference answers alone into the sole reward for on-policy post-training of reasoning LLMs. We call this method Reinforced Adversarial Learning (RAL). We first validate RAL on three verifiable tasks (integer multiplication, code-based arithmetic-expression synthesis, and SMILES-to-formula translation), using the verifiers only for evaluation. With sufficiently capable judges and enough training, RAL matches direct RLVR. We then apply RAL to scientific abstract generation in two settings. In the first setting, RAL guides the generator to produce abstracts in the scientific domain of the reference abstracts, even though the prompt leaves that domain unspecified. In the second setting, we generate abstracts for U.S. National Institutes of Health grant proposals conditioned on the grant title. Human scientists and an ensemble of LLM judges find our generations harder to distinguish from real abstracts than those of frontier-LLM baselines. These results suggest a path toward post-training reasoning models in scientific and other high-value domains, where increasingly capable judges infer hard-to-specify reward functions from example answers.
}

\ificlrstyle

\title{\papertitle}
\author{%
Albert Bou$^{*}$, Javier Valdes-Aleman, Jon M. Laurent, \\
Samuel G. Rodriques, Andrew D. White, Michaela Hinks \\
\& Siddharth Narayanan$^{*}$ \\
Edison Scientific Inc. \\
San Francisco, CA, USA \\
$^{*}$Correspondence to \texttt{abou@edisonscientific.com, sid@edisonscientific.com}
}
\maketitle
\begin{abstract}
\abstracttext
\end{abstract}
\else

\begin{center}
\begin{headerbox}
  \vspace*{1em}
    {\sffamily \textbf{\huge \papertitle}}

    \vspace{0.6em}
    {\sffamily \large \papersubtitle}

    \vspace{1em}
    {Albert Bou$^{*}$},
    {Javier Valdes-Aleman},
    {Jon M. Laurent},
    {Samuel G. Rodriques},
    {Andrew D. White},
    {Michaela Hinks},
    {Siddharth Narayanan$^{*}$}

    \vspace{1em}

    \sffamily \textbf{\large Abstract}

    \vspace{0.2em}

    \abstracttext
    \\

    \vspace{1em}

    {\footnotesize
    \raggedright
    Edison Scientific Inc., San Francisco, CA, USA\\
    $^{*}$Correspondence to \texttt{abou@edisonscientific.com, sid@edisonscientific.com}\\
    }
    \vspace{-1cm}\hfill\raisebox{0cm}{\includegraphics[width=1.3cm]{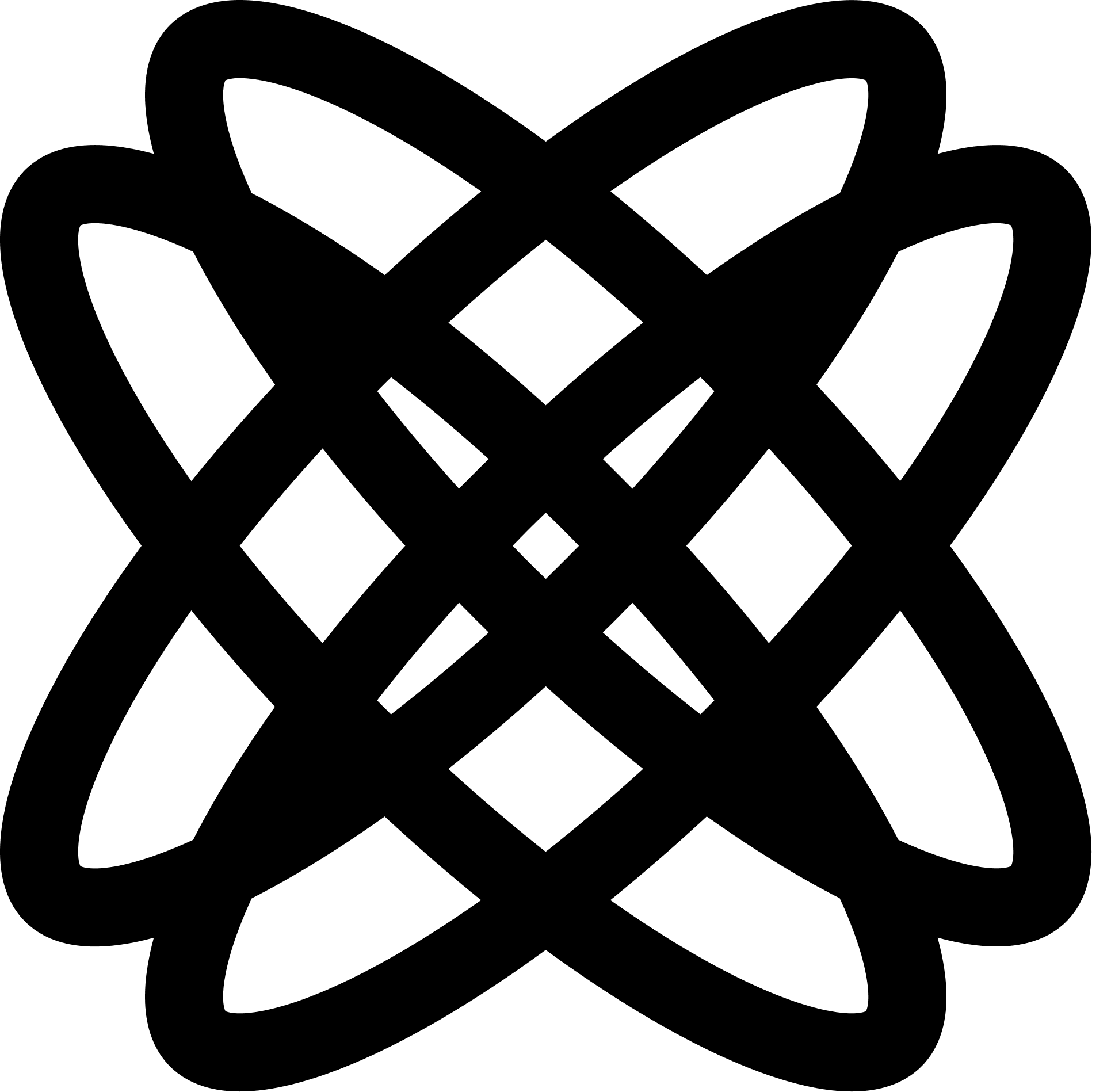}}\hspace{0cm}

\end{headerbox}
\end{center}
\fi

%% file: sections/01-main/sec-01-introduction.tex
\section{Introduction}
\label{sec:intro}

Many high-value domains for modern LLM post-training share a common asymmetry: high-quality answers are easy to recognize, and perhaps easy to collect, yet difficult to score with a reliable reward. A well-designed experimental protocol, a useful literature review, or a convincing interpretation of experimental results may be recognizable to an expert at a glance, but is difficult to capture in a hand-designed reward that accounts for all the relevant nuances. Reinforcement learning with verifiable rewards (RLVR), the engine behind recent reasoning advances in LLMs~\citep{shao2024deepseekmath, guo2025deepseek}, is limited to settings where a checker exists. LLM judges with frozen weights can score a broader range of outputs using a rubric~\citep{zheng2023judging}, but effective rubrics can be challenging to design and often require domain expertise. Their reward signals can also become brittle and exploitable under direct optimization. 

Recent adversarial approaches bring demonstration-based rewards to modern LLM post-training. However, they require explicit verifiers or demonstration reasoning traces, focus on teacher distillation, or fall short of RLVR~\citep{ye2025black,cai2025escaping,fanconi2025learning,damani2026right}.
We ask whether a sufficiently capable judge, independent from the generator and trained in parallel with it, can turn reference answers alone into an effective continuous reward for reasoning model post-training, yielding performance comparable to RLVR. 
The training loop is simple: A judge is trained to estimate the probability that an output comes from the reference set rather than the current generator, while the generator is trained with RL to produce outputs to which the judge assigns a high reference probability. The references do not contain reasoning traces or process supervision. As in RLVR, useful reasoning must therefore emerge from outcome-level feedback. Unlike RLVR, however, that feedback is learned by the judge from reference answers rather than supplied by a verifier. 
As the generator improves, the judge’s decision boundary moves with it. A natural expectation is that this game remains superficial: the judge separates the two distributions using cues such as length, formatting, or lexical choice, the generator learns to exploit those cues, and neither is pushed toward the properties that make a reference output good. However, with sufficiently capable models, the adversarial game can act as a curriculum: as the generator matches superficial features of the reference distribution, those features become less informative, pushing the judge toward progressively deeper properties. How far this process can go depends on the difficulty of inferring those properties from examples and on the judge’s capacity. Our results suggest that today’s LLM judges can move beyond superficial distinctions and, with sufficient co-training, provide an effective training signal. We call our approach Reinforced Adversarial Learning (RAL). We implement RAL on top of Prime-RL~\citep{primeintellect2025prime-rl} and release the code at
\ifanon
\else
\url{https://github.com/EdisonScientific/prime-rl-adversarial-paper}
\fi
.

\ifanon
\else
\begin{figure}[t]
    \centering
    \includegraphics[width=\linewidth]{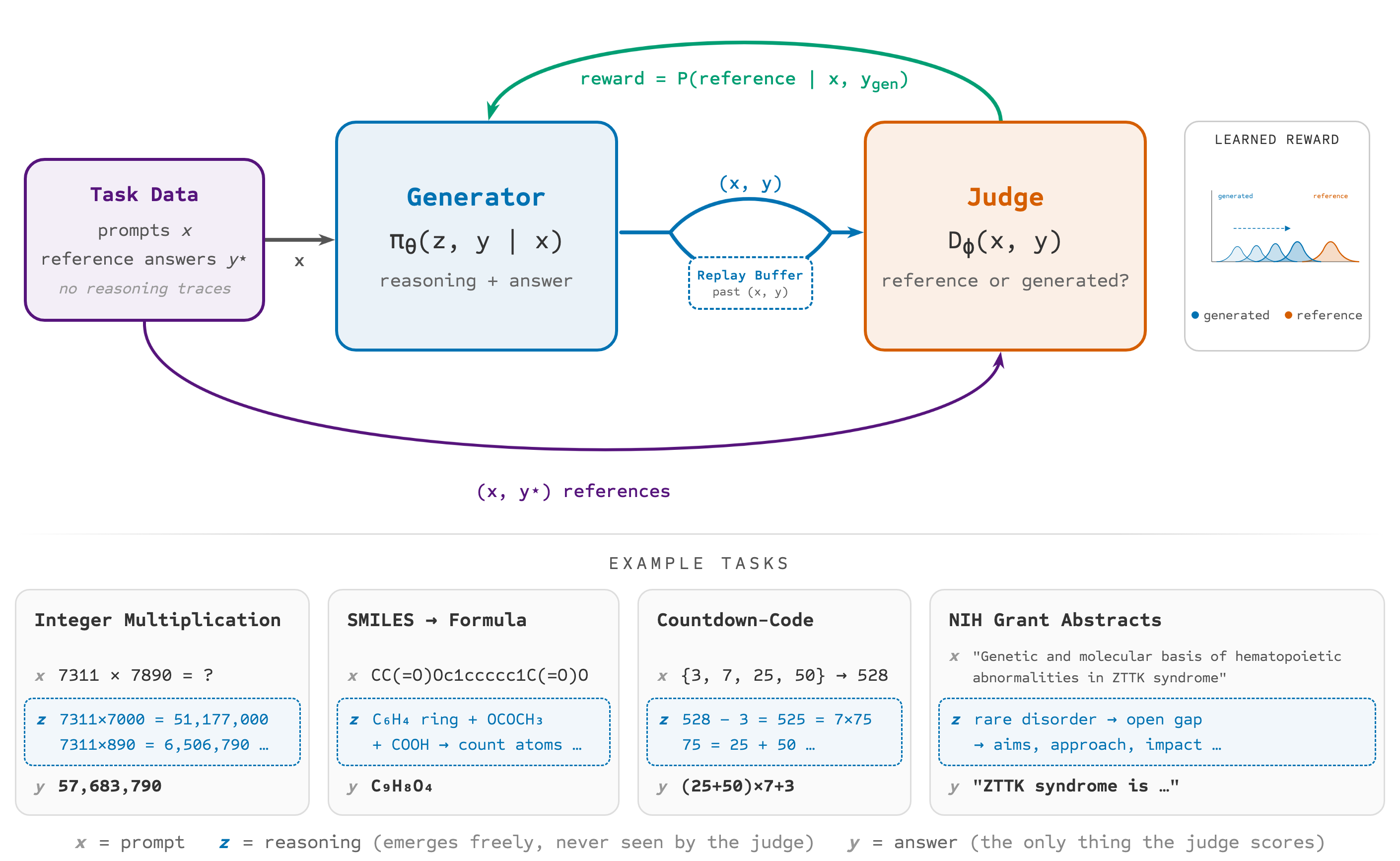}
    \caption{\textbf{Reinforced Adversarial Learning (RAL).}
A co-trained judge learns to tell reference answers from generated ones, and its reference
probability is the generator's only reward. With no verifiers or reasoning demonstrations,
reasoning $z$ emerges from answer-level feedback. Bottom: example tasks, with illustrative reasoning.}
    \label{fig:ral-schematic}
\end{figure}
\fi

We test this hypothesis across five settings spanning verifiable reasoning tasks and open-ended scientific text generation. In the first three tasks, ground-truth correctness exists but is withheld from training. On integer multiplication and code-based arithmetic-expression synthesis, judge-only training improves held-out correctness and can match or surpass training with exact rewards, including exact rewards augmented with a generator-sized judge. The same is observed when training a model to convert SMILES strings to chemical formulae. Because expected answers (numbers, expressions, or formulas) leave little room for stylistic shortcuts, these gains suggest that LLM judges can recover task-relevant structure using only comparisons between reference and generated outputs. We next study unconstrained scientific paper abstract generation using reference examples drawn exclusively from medical and biomedical domains. Our approach learns to produce abstracts in these domains without being told the target domain, while preserving semantic diversity. Finally, we turn to abstract generation for U.S. National Institutes of Health (NIH) grant applications, a practical setting where high-quality examples are available but no reliable verifier can be constructed. Starting from only a title and a prompt asking for a plausible grant abstract, our approach produces outputs that better match the NIH reference distribution than those of strong frontier-LLM baselines and are harder for both LLM judges and human evaluators to distinguish from real abstracts under the same evaluation protocol. Across these experiments, we observe evidence consistent with the hypothesized curriculum: stronger judges learn better reward signals, and longer co-training yields further gains in generator performance.


\ifanon
\else
Our contributions are as follows. First, we show that a simple judge trained only to estimate the probability that an individual answer comes from the reference distribution can provide the sole reward for reasoning LLM post-training, matching or exceeding explicit verifiable rewards across arithmetic, code, and chemistry tasks. These results support the view that, with modern LLMs, the adversarial game can move beyond superficial differences. Second, we identify judge capacity and continued co-training as key ingredients that make this simple adversarial signal effective. Third, we show that the same recipe extends to open-ended scientific generation, moving the policy model toward the reference distribution while preserving diversity and, on NIH grant abstracts, outperforming strong frontier-LLM baselines in fooling both LLM judges and human experts. Together, these results point to a broader post-training paradigm in which rewards are learned from reference answers alone, enabling to train reasoning models in domains where good answers are available but reliable verifiers, reward labels, and reasoning traces are not.
\fi 


%% file: sections/01-main/sec-02-related-work.tex
\section{Related Work}
\label{sec:related_work}


\paragraph{Inverse reinforcement learning.}
Inverse reinforcement learning (IRL) learns rewards from expert demonstrations rather than requiring explicitly specified rewards~\citep{ng2000algorithms}. GAIL and AIRL established adversarial learning from demonstrations in control settings~\citep{ho2016generative,fu2017learning}. Early language applications adapted these ideas to recurrent models and smaller pretrained language models, before reasoning-oriented post-training became widespread in larger LLMs~\citep{shi2018toward,li2019dialogue,wu2021textgail}. TextGAIL is structurally similar to RAL but evaluates text quality and diversity, without investigating whether learned rewards can recover the structure of an underlying exact reward and guide the generator toward exact answers. Furthermore, TextGAIL trains the generator on expert trajectories, which would require expert reasoning traces when applied to modern reasoning models. RAL uses reference answers only to train the judge, leaving the generator to learn its own reasoning from outcome-level rewards.

\paragraph{Reward models for LLM post-training.}
Reinforcement learning from human feedback established learned outcome reward models (ORMs) as a post-training signal, assigning a scalar to a completed response~\citep{christiano2017deep,stiennon2020learning,ouyang2022training}. Because static ORMs are vulnerable to over-optimization and reward hacking~\citep{gao2023scaling}, recent reasoning pipelines instead use verifiable rewards (RLVR), replacing the learned model with rule-based correctness checks~\citep{shao2024deepseekmath,guo2025deepseek}. Process reward models (PRMs) provide step-level feedback~\citep{uesato2022solving,lightman2024let,wang2024math}, but require costly process annotations or, in the case of implicit PRMs, at least outcome-level verification~\citep{yuan1981free}. These approaches require explicit quality supervision from human labels, a rubric, or a verifier, whereas RAL targets settings where high-quality outputs are easy to collect but hard to score.

\paragraph{Co-training reward models and policies.}
Co-training offers another response to the limitations of static reward models. As the policy improves, its outputs can shift beyond a fixed reward model's training distribution. Several methods therefore update the reward model alongside the policy. PRIME~\citep{cui2025process} co-trains an implicit PRM, while TANGO~\citep{zha2026rl} trains a generative process verifier with RL alongside the generator. AIRL-S~\citep{jin2025your} co-trains an adversarial process reward model using correctness-filtered reasoning trajectories and combines its rewards with a binary outcome reward. All three rely on explicit outcome correctness supervision. In contrast, RAL co-trains the judge and policy using reference answers, without a task verifier or correctness labels for generated outputs.

\paragraph{Adversarial LLM post-training from demonstrations.}
Several recent approaches combine reward–policy co-training with adversarial learning from demonstrations, but differ in their supervision requirements and training procedures. GAD~\citep{ye2025black} distills a stronger teacher through a co-trained discriminator that distinguishes teacher from student responses. R-AIRL~\citep{fanconi2025learning} avoids verifiers but requires expert reasoning traces, constructing process supervision from expert answers and artificially corrupted traces. VARL~\citep{damani2026right} trains a pointwise discriminator but multiplies its reward by the output of an explicit verifier, which still determines task success. RARO~\citep{cai2025escaping} learns from demonstrations alone, training a single model for  both generation and discrimination with GRPO. It derives rewards from three possible verdicts in pairwise comparisons of reference and policy outputs (reference wins, policy wins, or tie), and its reported accuracy remains below RLVR. RAL instead learns a continuous reward from reference answers alone, without requiring verifiers, expert reasoning traces, or separately constructed negatives. It co-trains a separate, independently scalable judge to estimate the probability that an answer comes from the reference set rather than the current policy. The judge’s reference probability serves as the sole RL reward, allowing reasoning to emerge from outcome-level feedback without reasoning demonstrations during post-training. RAL achieves performance comparable to RLVR in our experiments.


%% file: sections/01-main/sec-03-method.tex
\section{Method}
\label{sec:method}

We now formalize RAL in the verifier-free setting, where reference answers provide the only task supervision. We assume a dataset $\mathcal{D}={(x_j,y_j^\star)}_{j=1}^N$, where $x_j \in \mathcal{X}$ is a task input prompt and $y_j^\star \in \mathcal{Y}$ is a reference final answer drawn from a reference distribution $p_{\mathrm{ref}}(\cdot \mid x_j)$. We do not assume access to a reward function or verifier that can score arbitrary generations.

Following adversarial imitation learning~\citep{ho2016generative,fu2017learning}, RAL jointly trains a generator and a judge that distinguishes reference answers from generated answers. Both are LLMs. The generator defines a policy $\pi_\theta(z,y \mid x)$ over reasoning traces $z$ and final answers $y$. The judge $D_\phi(x,y) \in [0,1]$ estimates the probability that an answer comes from the reference distribution rather than the generator. It receives only $x$ and $y$, never $z$, during both training and reward evaluation. The judge outputs a single constrained token, either \texttt{A} or \texttt{B}, corresponding to reference and generated answers, respectively. We then define $D_\phi(x,y) = P_\phi(\texttt{A}\mid x, y)$. 

For a sampled task prompt $x_j$ with reference answer $y_j^\star$, the generator samples a reasoning trace and final answer $(z_{j,i},y_{j,i}) \sim \pi_\theta(\cdot \mid x_j)$. Malformed rollouts that do not follow formatting instructions receive zero reward and are excluded from judge training, to prevent reasoning traces leaking into judge inputs. Each generator sample contributes its corresponding reference $(x_j,y_j^\star)$ to the judge training batch, while only valid generations $(x_j,y_{j,i})$ contribute generated examples. The judge batch therefore contains $1-2\times$ as many examples as the generator batch. The judge is trained with a cross-entropy loss. The generator is updated on the full rollout $\mathrm{concat}(z_{j,i}, y_{j,i})$ using the judge's reference probability as reward. We use GRPO~\citep{shao2024deepseekmath}, although RAL can in principle use other on-policy RL objectives. Judge training uses samples from the current generator in the basic setup, but its supervised objective also permits replayed or off-policy generations. The generator's reward is therefore $r_{\mathrm{RAL}}(x,z,y) = \mathbf{1}[\mathrm{valid}(z,y)]\,D_\phi(x,y)$, where $\mathrm{valid}(z,y)$ denotes task-specific format parsing of the full rollout (typically requiring \texttt{<think>} and \texttt{<answer>} tags). 
We chose to use the judge probability as the reward empirically for stability. Recent work provides complementary theoretical motivation for this choice. In the verifier-gated setting of VARL~\citep{damani2026right}, using the discriminator probability directly yields bounded per-sample rewards and, under an optimal discriminator, makes reward maximization equivalent up to an affine transformation to minimizing the Vincze--Le Cam divergence. The logit transformation, by contrast, is unbounded and can lead to high-variance policy-gradient estimates. Unlike VARL, RAL uses the judge as the sole training reward rather than as an auxiliary signal combined with an explicit verifier.

GRPO samples a group of $G$ outputs for each input and computes each output's advantage by subtracting the group's mean reward from its reward and dividing by the group's reward standard deviation~\citep{shao2024deepseekmath}. Malformed outputs thus incur a negative advantage whenever other outputs in their group earn reward. We also apply a KL penalty toward the initial policy for regularization. To stabilize co-training, we use a \emph{judge update gate}. After a warm-up period, judge updates are skipped when an exponential moving average of classification accuracy on the current judge batch, measured before the update, exceeds a threshold. Otherwise, the judge is updated on the current batch. This is intended to keep the judge from becoming too strong too quickly and to preserve a useful learning signal for the generator. A maximum skip streak ensures that the judge periodically updates against the current generator distribution.

We use two optional stabilization techniques in the final experiment. First, we augment each judge batch with randomly sampled past generations from a replay buffer to reduce forgetting of earlier generator strategies. Second, we use soft updates to smooth changes in the reward. We maintain two copies of the judge: one is trained on classification batches, and the other computes rewards. The reward copy tracks the training copy through Polyak averaging~\citep{lillicrap2015continuous}, $\bar{\phi} \leftarrow (1-\tau)\bar{\phi}+\tau\phi$, where $\phi$ denotes the training weights, $\bar{\phi}$ the reward weights, and $\tau \in (0,1]$ the averaging coefficient. Rewards are then computed with $D_{\bar{\phi}}(x,y)$. This is straightforward to implement in post-training frameworks like Prime-RL\citep{primeintellect2025prime-rl} that already separate training from inference: when synchronizing the two copies, we blend the new training weights with the existing inference weights instead of replacing them outright.

%% file: sections/01-main/sec-04-experiments.tex
\section{Experiments}
\label{sec}
\begin{figure}
    \centering
    \ifanon
\else
        \includegraphics[width=\linewidth]{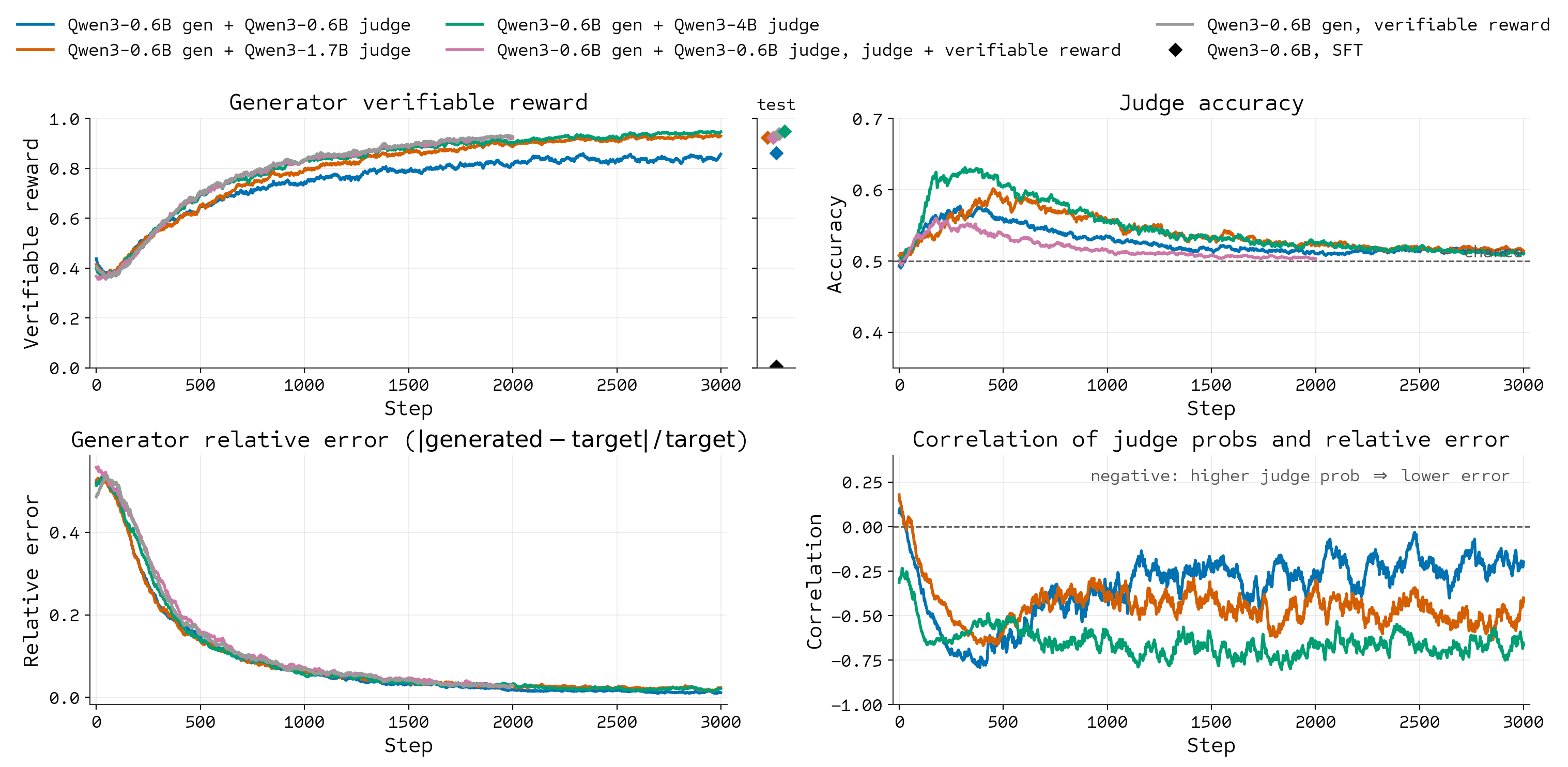}
    \fi
    \caption{\textbf{RAL with 1.7B and 4B judges matches verifiable-reward training on integer multiplication.}
Generator accuracy, judge accuracy, \anontext{}{relative error} and judge-score correlation with relative error. Diamonds denote held-out test performance, including SFT. Y-axes are truncated.}
    \label{fig:digit-multiplication-results}
\end{figure}

We train and evaluate RAL on five settings spanning multiplication, code-based
arithmetic-expression synthesis, chemistry, and scientific text generation. All generators and judges are drawn from the Qwen3 family of models~\citep{yang2025qwen3}. Generators produce reasoning traces before their final answers, while judges are constrained to output a single classification token, \texttt{A} or \texttt{B}, without generating reasoning traces. To isolate the effect of judge capacity, we hold the generator and training hyperparameters fixed and vary only the judge size. We extend the co-training horizon to study the effect of additional training. Training hyperparameters are provided in Appendix~\ref{app:hyperparameters}.

\paragraph{Task suite.}

We begin with three tasks where verifiable rewards are withheld from RAL training and used only for evaluation, allowing us to measure how well a judge can recover the true task objective from examples alone and use it to train the generator. In integer multiplication, the model is given two four-digit integers and must produce their exact product. In Countdown-Code~\citep{damani2026right}, the model is given a set of integers and a target value. It must then edit a small Python solution file by assigning an arithmetic expression that uses each integer exactly once and the operations $+$, $-$, $\times$, and $/$ to reach the target. The submission is checked by an accompanying Python test, and we separately evaluate whether the expression itself satisfies the original arithmetic constraint. The test file is also editable, allowing us to detect if the model obtains reward by modifying the test rather than solving the intended task. In SMILES-to-formula prediction, the model is given a molecule as a SMILES~\citep{weininger1988smiles} string and must derive its exact molecular formula. The constrained answer formats of these tasks leave limited room for stylistic variation, making held-out correctness a useful test of whether the judge learns task-relevant structure.

We next move to scientific writing, where outputs are open-ended and no reliable verifier is available. We first study unconstrained research-paper abstract generation. The policy model is trained with the same generic instruction for every example, asking it to produce a credible research-paper abstract. The reference set only consists of biomedical abstracts from paper in these domains, but the model is not told the target domain in the prompt. The judge must infer the target topic distribution from the reference examples and convey it to the generator through its reward. This lets us test whether RAL can move generation toward the target distribution without collapsing diversity. 

Finally, we study NIH grant abstract generation, a realistic title-conditioned scientific writing task. Each input is the title of an NIH grant proposal (encoding a narrow topic), and the model must generate a plausible abstract for that proposal. References are drawn from NIH grants funded in FY2026, to reduce likelihood of leakage into base LLM pre- and post-training data. Unlike the previous abstract experiment, the target distribution is conditioned on the proposal title, requiring the judge to learn both what NIH abstracts look like and whether an abstract fits a specific proposal title. 
Full task and evaluation details for all five settings are provided in Appendix~\ref{app:tasks}.

\begin{figure}
    \centering
    \ifanon
\else
        \includegraphics[width=\linewidth]{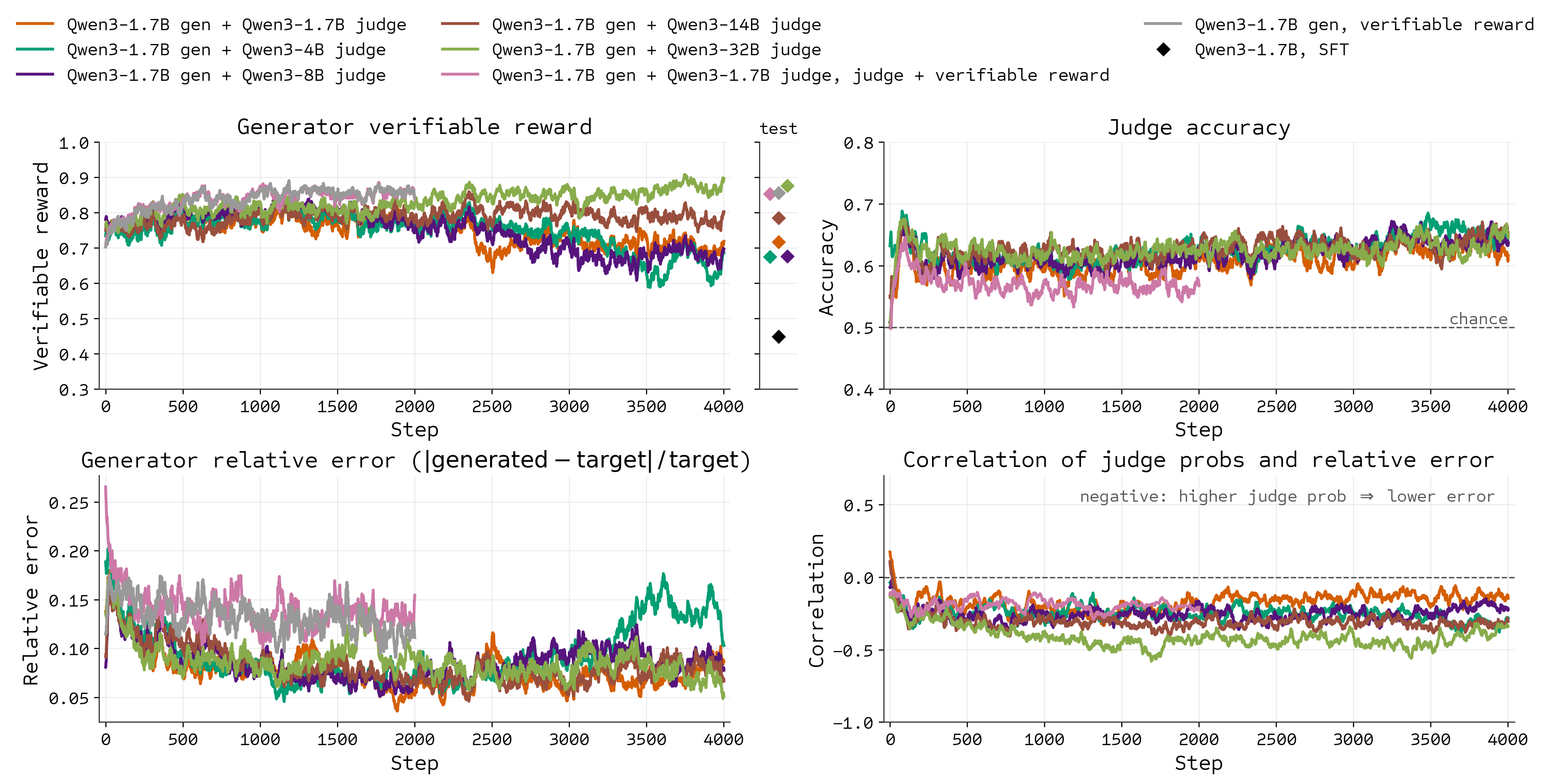}
    \fi
    \caption{\textbf{Judge capacity and co-training time drive performance on Countdown-Code.}
Generator verifiable reward, judge accuracy, \anontext{}{generator relative error} and correlation between judge score and relative error. Diamonds denote held-out test performance at each run's cutoff, including SFT ($\sim 45\%$). The 32B judge surpasses both verifiable-reward baselines after 4{,}000 training steps using reference answers alone. Y-axes are truncated for readability.}
    \label{fig:countdown-results}
\end{figure}

\paragraph{Baselines.}
For the three tasks with held-out verifiable rewards, we compare RAL with three diagnostic baselines and a judge-capacity ablation. The \emph{verifiable reward} baseline uses only the task evaluator as the generator reward  (RLVR). The \emph{judge + verifiable reward} baseline multiplies the judge reward by the verifiable reward (as in VARL). The \emph{supervised fine-tuning (SFT)} baseline trains directly on the reference examples until convergence. The ablation uses RAL with a judge of the same capacity as the generator. The judge + verifiable reward baseline uses the same capacity setting, following prior adversarial approaches~\citep{ye2025black,cai2025escaping,damani2026right}. We evaluate each model regularly during training and report the best accuracy achieved on the held-out evaluation set. Because RAL must learn its reward online, we train it beyond the baseline horizon to test whether additional co-training closes the gap to direct verifiable-reward optimization. For biomedical-paper abstract generation, we compare RAL with optimization against a frozen Haiku judge. For NIH grant abstract generation, we compare RAL with frontier LLMs and an SFT baseline trained on the reference abstracts, evaluating all generated abstracts against real NIH abstracts.

\paragraph{Scaling.}
We study the effect of scaling both the generator and judge models independently, unlike prior work that keeps them coupled~\citep{damani2026right,wu2021textgail}. We hypothesize that learning complex task structures and reference distributions is limited by judge capacity more than the generator, and therefore study both scaling dimensions.

\subsection{Verifier-held-out tasks results}

\paragraph{Integer multiplication.}
On the four-digit integer multiplication task, RAL learns an effective reward from reference answers alone, guiding the generator toward correct solutions. As shown in~\autoref{fig:digit-multiplication-results} (\anontext{left}{top left}), the 0.6B judge drives the generator to approximately 85\% accuracy. Larger judges learn effective rewards faster. The 1.7B and 4B judges nearly match the \emph{verifiable reward} and \emph{judge + verifiable reward} baselines by 2{,}000 steps and close the remaining gap by 3{,}000 steps. In contrast, supervised fine-tuning on the same references reaches only $\sim 0.5\%$ held-out accuracy (gray diamond in the test column), showing that direct imitation of reference answers is insufficient in this setting.

Relative error falls across all judge sizes (\anontext{see Appendix section \ref{app:mult-relerr}}{bottom left}), mirroring the increase in accuracy. 
The negative correlation between judge reward and relative error is stronger for larger judges, particularly later in training (\anontext{right}{bottom right}). This suggests that larger judges learn rewards better aligned with the hidden verifiable objective, which may explain their faster gains in correctness. As the generator improves, judge accuracy approaches chance at 0.5 (\anontext{middle}{top right}), while judge scores remain correlated with lower error. Manual inspection also suggests that the learned reasoning traces are comparable to those produced under verifiable-reward training.

\begin{figure}
    \centering
    \ifanon
\else
        \includegraphics[width=\linewidth]{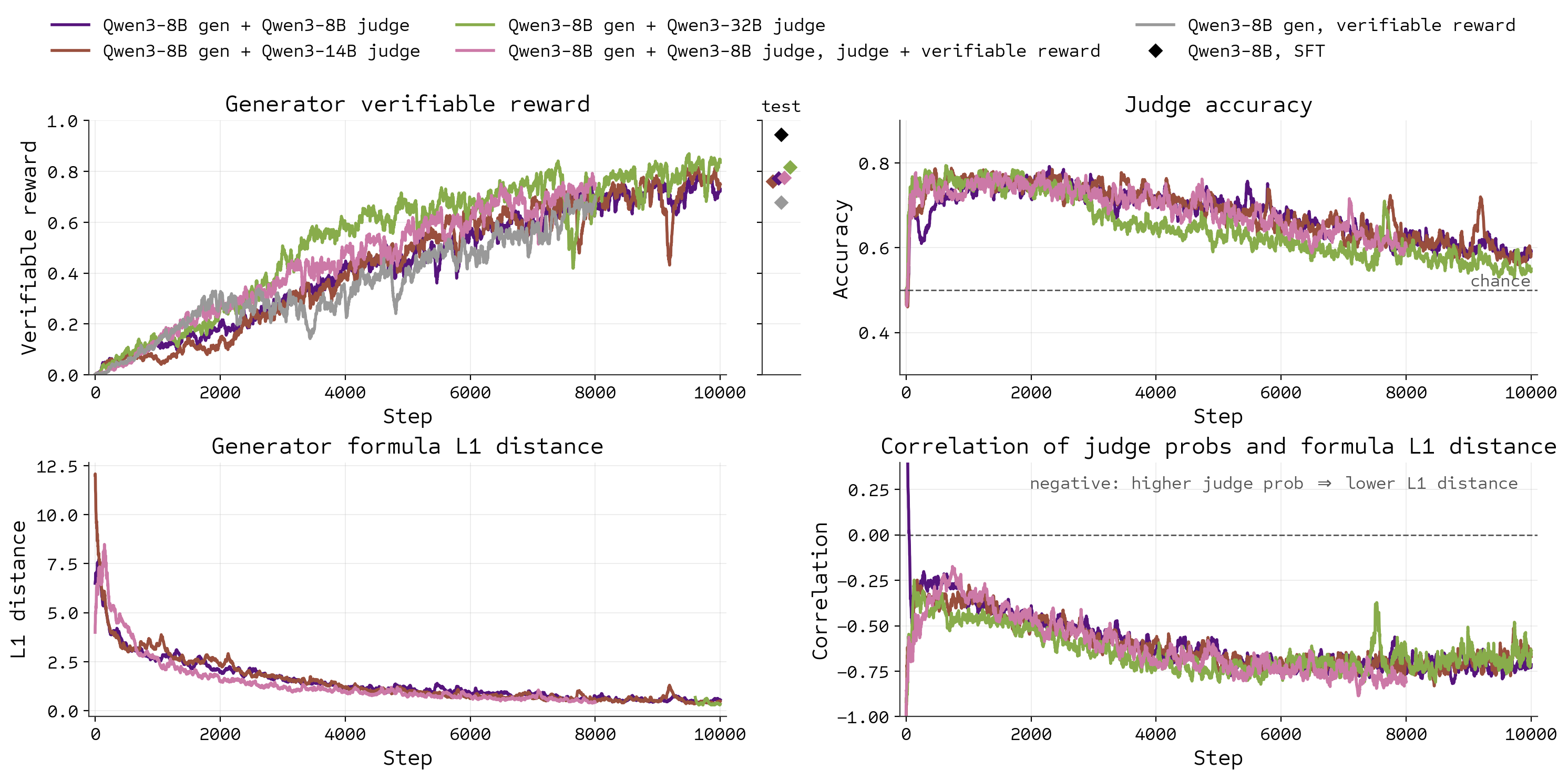}
    \fi
    \caption{\textbf{RAL with a 32B judge outperforms binary verifiable-reward training on SMILES-to-formula.}
Generator verifiable reward, judge accuracy, \anontext{}{generated formula L1 distance to target} and judge-score correlation with formula L1 distance. Diamonds show held-out test performance at each run's cutoff. Y-axes are truncated.}
    \label{fig:smiles2formula-results}
\end{figure}

\paragraph{Countdown-Code.}
The Countdown-Code task tests RAL on a more demanding structured reasoning task. As in four-digit multiplication, results on Countdown-Code confirm the importance of both judge capacity and co-training time (\autoref{fig:countdown-results}). Smaller judges plateau or degrade, while larger judges support continued improvement. The 14B judge approaches the verifiable-reward baselines, and the 32B judge surpasses them by 4{,}000 steps. By contrast, supervised fine-tuning on the same references reaches only $\sim 45\%$ held-out reward (gray diamond, \anontext{left}{top left}). This gap shows that direct imitation alone is insufficient in this setting and is consistent with RAL eliciting useful reasoning through outcome-level feedback.

RAL also drives relative error more steadily toward zero, while the verifier-based baselines retain higher error despite strong verifiable reward (\anontext{see Appendix section \ref{app:countdown-relerr}}{bottom left}). This suggests that the learned judge reward captures how close incorrect solutions are to the target, providing a useful signal beyond binary task success. Larger judges show a stronger negative correlation between reward and relative error (\anontext{right}{bottom right}), providing further evidence that capable judges can recover task-relevant structure from reference answers without access to the verifier. Judge accuracy remains above chance throughout training. This may partly reflect differences between generated and reference solutions, since Countdown admits multiple correct answers, though further training may narrow this gap. Even so, the judge’s reward remains informative, guiding the generator toward more accurate solutions. We observe little reward hacking through test-file modification, unlike the findings of~\cite{damani2026right}, whose setup includes pretraining on cheating trajectories to induce hacking.

\paragraph{SMILES-to-formula.}
In this setting, we test RAL on a chemistry task that requires recovering molecular composition from a structured chemical representation. As in the preceding tasks, RAL improves accuracy and learns rewards that track task-relevant errors without access to the evaluator (Figure~\ref{fig:smiles2formula-results}). Here, the distinctive result is that RAL, particularly with the 32B judge, improves accuracy faster than training with binary verifiable rewards (\anontext{left}{top left}). One possible explanation is that binary rewards yield zero advantage whenever a prompt's rollouts are all incorrect or all correct, whereas the judge can assign different scores to these answers and provide a learning signal. We observe nonzero reward advantages on substantially more rollout groups under RAL. This feedback may be particularly useful here because the generator starts at lower accuracy than on the preceding tasks, making groups with no correct answers more common. Supervised fine-tuning reaches $\sim 94\%$ held-out reward on this task (grey diamond, \anontext{left}{top left}), suggesting that explicit reasoning may be less important here than in the preceding tasks.

\begin{figure}
    \centering
    \includegraphics[width=\linewidth]{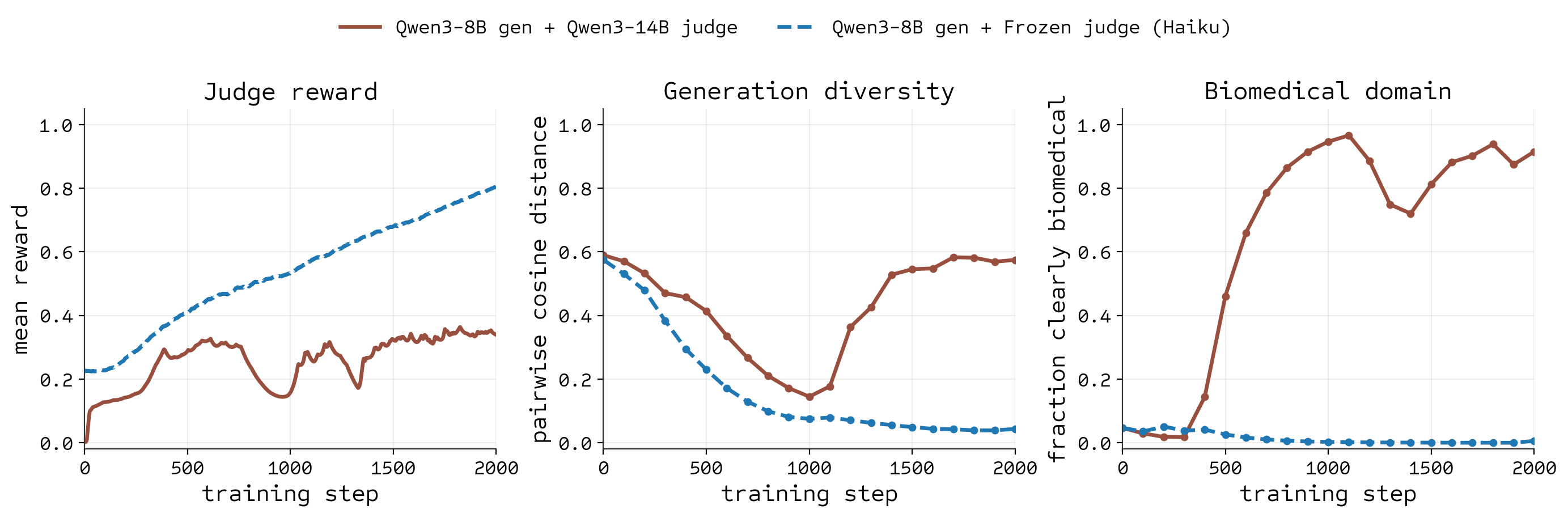}
    \caption{\textbf{Co-training recovers the hidden biomedical domain without collapsing diversity.} A frozen Haiku judge yields increasing reward but collapses generation onto a narrow, out-of-domain template. RAL uses a co-trained Qwen3-14B judge. Biomedical rate is assessed by an LLM classifier; diversity is mean pairwise cosine distance between text embeddings.}
    \label{fig:medical-abstracts-results}
\end{figure}

\subsection{Open-ended Paper Abstract Generation results}
\label{sec:medical_abstracts}

Open-ended paper abstract generation tests whether RAL can recover a hidden target distribution without collapsing diversity. The reference abstracts come exclusively from biomedical papers, but the generator does not receive explicit domain instructions. Because a frozen judge (with sufficient prompting) can in theory determine the domain and quality of an abstract, we compare RAL against rewards from a frozen judge (Claude Haiku 4.5~\cite{anthropic2025haiku45}). To make the discrimination task easier for Haiku, we present a generated abstract alongside a randomly sampled real reference abstract and ask it to identify the real one. The generator receives a binary reward of $1.0$ if Haiku selects the generated abstract as real, and $0.0$ otherwise. We also explicitly tell Haiku the target domain by including the following sentence in its prompt: ``The real paper corpus contains medical and biomedical paper abstracts.'' Thus, the frozen judge receives both a reference abstract for direct comparison and explicit domain information. The frozen-judge prompt is provided in Appendix~\ref{app:haiku_prompt}.

We evaluate whether generated abstracts match the target topic distribution and how semantically diverse they are. Domain alignment is measured as the fraction of abstracts classified as clearly biomedical by Claude Haiku 4.5 using a separate evaluation prompt (Appendix~\ref{app:haiku_prompt}). Semantic diversity is computed as the mean pairwise cosine distance between embeddings computed with OpenAI’s \texttt{text-embedding-3-small} model~\citep{openai2024embeddings}. Figure~\ref{fig:medical-abstracts-results} shows that optimization against the frozen judge steadily increases reward (left) while reducing semantic diversity (middle) and failing to recover the biomedical domain (right). At the final checkpoint, 125 of 128 generations collapse onto a Pacific-oyster microplastics ecotoxicology theme despite remaining lexically distinct (Appendix~\ref{app:external_judge_failure}). RAL instead drives the biomedical rate toward one while preserving diversity. Diversity initially falls, then recovers to approximately its starting level while domain fidelity remains high. These results show that RAL can recover the target domain from reference examples alone without collapsing onto a narrow topic.

During training, RAL generator reward and judge accuracy show a sawtooth pattern as the generator and judge adapt to each other. This pattern is consistent with the generator learning features that fool the current judge, raising reward, followed by the judge learning to distinguish the new generations, lowering reward and redirecting learning toward remaining differences from the references. We examine these dynamics in more detail in the final experiment.

\begin{table}[t]
\centering
\caption{RAL and baseline performance across tasks. Values are means with 95\% confidence intervals ($\pm$ truncated at 0).}
\label{tab:main-results}

\begin{minipage}[t]{0.48\linewidth}
\centering
\small
\textbf{Integer Multiplication}\par\smallskip
\begin{tabular*}{\linewidth}{@{\extracolsep{\fill}}lr@{}}
\toprule
Method & Accuracy \\
\midrule
Verifiable reward             & 0.938 {\scriptsize$\pm$0.017} \\
Judge + verifiable reward     & 0.925 {\scriptsize$\pm$0.018} \\
SFT                           & 0.005 {\scriptsize$\pm$0.008} \\
RAL (0.6B judge)              & 0.861 {\scriptsize$\pm$0.024} \\
RAL (4B judge)                & \textbf{0.949} {\scriptsize$\pm$0.016} \\
\bottomrule
\end{tabular*}
\end{minipage}
\hfill
\begin{minipage}[t]{0.48\linewidth}
\centering
\small
\textbf{Countdown-Code}\par\smallskip
\begin{tabular*}{\linewidth}{@{\extracolsep{\fill}}lr@{}}
\toprule
Method & Reward \\
\midrule
Verifiable reward             & 0.858 {\scriptsize$\pm$0.024} \\
Judge + verifiable reward     & 0.854 {\scriptsize$\pm$0.024} \\
SFT                           & 0.449 {\scriptsize$\pm$0.034} \\
RAL (1.7B judge)              & 0.718 {\scriptsize$\pm$0.031} \\
RAL (32B judge)               & \textbf{0.878} {\scriptsize$\pm$0.023} \\
\bottomrule
\end{tabular*}
\end{minipage}

\bigskip

\begin{minipage}[t]{0.48\linewidth}
\centering
\small
\textbf{SMILES-to-Formula}\par\smallskip
\begin{tabular*}{\linewidth}{@{\extracolsep{\fill}}lr@{}}
\toprule
Method & Accuracy \\
\midrule
Verifiable reward             & 0.679 {\scriptsize$\pm$0.033} \\
Judge + verifiable reward     & 0.776 {\scriptsize$\pm$0.029} \\
SFT                           & \textbf{0.945} {\scriptsize$\pm$0.016} \\
RAL (8B judge)                & 0.773 {\scriptsize$\pm$0.029} \\
RAL (32B judge)               & 0.816 {\scriptsize$\pm$0.027} \\
\bottomrule
\end{tabular*}
\end{minipage}
\hfill
\begin{minipage}[t]{0.48\linewidth}
\centering
\small
\textbf{NIH Grant Abstracts}\par\smallskip
\setlength{\tabcolsep}{3pt}
\begin{tabular*}{\linewidth}{@{\extracolsep{\fill}}lrr@{}}
\toprule
Method & LLM ensemble & Human \\
\midrule
SFT                & 0.04 {\scriptsize$\pm$0.04} & - \\
GPT-5.6 Sol        & 0.01 {\scriptsize$\pm$0.03} & 0.37 {\scriptsize$\pm$0.09} \\
Claude Opus 5      & 0.01 {\scriptsize$\pm$0.03} & 0.27 {\scriptsize$\pm$0.08} \\
Gemini 3.7 Flash   & 0.00 {\scriptsize$\pm$0.03} & 0.21 {\scriptsize$\pm$0.07} \\
RAL (235B judge)   & \textbf{0.35} {\scriptsize$\pm$0.08} & \textbf{0.55} {\scriptsize$\pm$0.09} \\
\bottomrule
\end{tabular*}
\end{minipage}
\end{table}

\subsection{NIH Grant Proposal Abstract Generation results}

\begin{figure}[!htb]
    \centering
    \ifanon
\else
        \includegraphics[width=\linewidth]{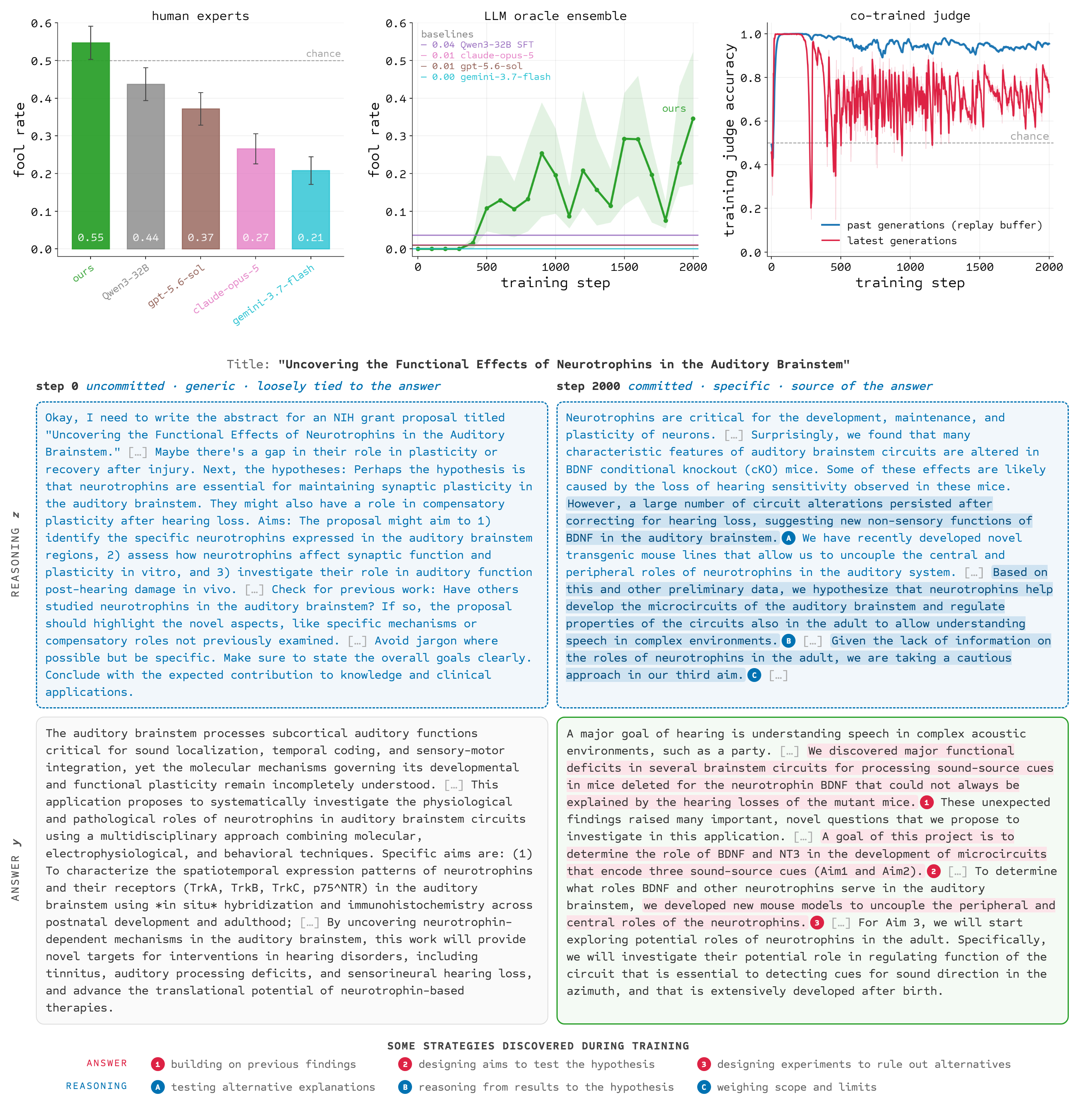}
        \caption{\textbf{RAL on NIH grant abstracts.}
        Top left: fooling rates against human experts for our model, the base model and frontier baselines. Error bars show ±1 SE, and the dashed line marks chance. Top middle: fooling rate against LLM judges over training. The curve shows the mean across five judges, with shading indicating their min–max range, and horizontal lines mark baseline rates. Top right: co-trained judge accuracy on current and replayed generations. Bottom: verbatim excerpts for one held-out title at steps 0 and 2000, with selected strategies highlighted in the answers and reasoning traces.}
    \fi
    \label{fig:nih-curriculum}
\end{figure}

NIH grant abstract generation is our most challenging verifier-free setting. We use Qwen3-32B as the generator. Earlier experiments show that performance improves with judge size, motivating our choice of Qwen3-235B-A22B, the largest model in the Qwen3 family, as the judge for this task. We include past generations from a replay buffer in judge training to help the judge remember features exploited by the generator in earlier iterations. To smooth changes in the reward, we use a Polyak-averaged copy of the judge for reward scoring.

We evaluate RAL and all baselines with five LLM judges (GPT-5.6 Sol~\citep{openai2026gpt56}, Claude Opus 4.8~\citep{anthropic2026opus48}, Claude Opus 5~\citep{anthropic2026opus5}, Gemini 3.1 Pro Preview~\citep{google2026gemini31pro}, and Gemini 3.7 Flash~\citep{google2026gemini37flash}) and with human experts. Each evaluator sees the proposal title alongside one real and one generated abstract and is prompted to identify the generated abstract.  We define the fooling rate as the fraction of comparisons in which the evaluator selects the real abstract as generated. To reduce sensitivity to any single judge’s preferences or blind spots, we report the mean fooling rate across the five LLM judges. All human evaluators hold or are pursuing a PhD in a biomedical field. Each human evaluator receives shuffled batches containing abstracts from all evaluated models.

\autoref{fig:nih-curriculum} shows that RAL substantially increases evaluator fooling rates. The generator before RAL training, frontier model baselines, and SFT achieve mean fooling rates of at most 4\% against the five LLM judges (top middle). RAL surpasses these baselines within 500 steps and reaches a mean fooling rate of approximately 35\% at the final checkpoint. Human evaluation (top left) shows the same preference for RAL, but with higher absolute fooling rates: RAL achieves a 55\% fooling rate, compared with 44\% for the generator before training and 21--37\% for the evaluated frontier models.

The training judge's accuracy on current generations repeatedly falls and recovers, while its accuracy on replayed generations remains high (top right). These cycles are consistent with a moving adversarial curriculum: new output features briefly fool the judge, then become less effective at fooling it as the judge adapts, even as the generator continues to use them. Sustained accuracy on replayed samples suggests that the judge retains the ability to recognize earlier generation strategies. Alongside these judge dynamics, qualitative inspection suggests that the generated abstracts develop more coherent scientific arguments. Later outputs include hypotheses motivated by purported preliminary findings, aims designed to test those hypotheses, and experiments that address alternative explanations. Qualitative inspection also suggests that reasoning traces become more specific and more closely reflected in the final abstracts. Early traces are generic and tentative, while later traces reason from those findings to hypotheses, consider alternative explanations, and weigh the scope and limits of the proposed work. These changes occur even though the judge never sees the traces and they receive no direct supervision.

These observations suggest that co-training helps the generator to progressively capture the underlying structure of reference abstracts, including how they connect problems, mechanisms, and proposed experiments. Our qualitative analysis covers only a subset of generations and does not capture the full range of patterns learned during training, particularly subtler distinctions. LLM judges may detect subtle differences between generated and reference abstracts that human evaluators miss, potentially explaining their lower fooling rates, though we do not test this explanation directly.

%% file: sections/01-main/sec-05-limitations-and-future-work.tex
\section{Limitations and future work}
\label{sec:limitations}
Our study has several limitations. To isolate judge capacity and co-training duration, we keep most other hyperparameters fixed across settings and do not perform extensive task-specific sweeps. Task (and judge size) specific tuning may improve both convergence and final performance. 

On the NIH grant experiment, RAL faces stability challenges associated with adversarial co-training. Despite the fluctuations in fooling rate seen in~\autoref{fig:nih-curriculum} (left), RAL repeatedly recovers from performance drops and reaches fooling rates above the frontier-model baselines at multiple checkpoints, leaving room to improve stability while preserving these gains. Furthermore, the generator is not currently conditioned on specific research results, only on the title. Conditioning it on real results and providing those results to the judge is a natural next step. 

Judge capacity is another practical bottleneck. Larger judges consistently learn better rewards, and some tasks require judges substantially larger than the generator to match verifiable-reward training. This increases training cost, though the judge can be discarded at inference time. Better judge initialization, specialized pretraining, or more efficient judge architectures may reduce this requirement. Continued improvements in model capability and inference efficiency may also make large judges increasingly practical, easing this bottleneck.

%% file: sections/01-main/sec-06-conclusion.tex
\section{Conclusion}
\label{sec:conclusion}

Our work shows that reference final answers alone can support effective RL post-training of reasoning models. RAL uses a simple supervised judge, trained alongside the generator but scaled independently, to turn reference data into an adaptive training objective. The judge’s predicted reference probability serves as the sole reward, without verifiers, reward labels, or expert reasoning traces. Useful reasoning is therefore learned through outcome-level feedback. On tasks with hidden verifiable rewards, this learned signal can match or exceed direct reward optimization. On open-ended scientific generation, RAL recovers a target domain from reference examples without explicit domain instructions, while preserving diversity. On NIH grant generation, it surpasses strong frontier-LLM baselines in fooling both LLM judges and human experts. Across settings, judge capacity and sufficient co-training are key to making this simple reward signal effective. These results suggest that modern LLM judges can recover task-relevant structure from reference answers and support the view that adversarial co-training can provide a curriculum that moves beyond superficial cues. The results across arithmetic, code, chemistry, and scientific writing motivate further study of this approach in scientific and other high-value domains where good answers are easier to collect than to score.

%% file: sections/01-main/sec-07-statement.tex
\section*{AI use statement}

In this work, we used generative AI tools to help format and clean datasets, suggest code changes to the Prime-RL training framework to test our ideas, suggest relevant related papers, and support qualitative analysis of model traces.
We used AI to generate synthetic data for the Countdown-Code task. This synthetic data was only used as part of the training mix; all reported performance metrics were computed with a deterministic verifier. 
We reviewed all AI-assisted work: we manually validated a subsample of each dataset for quality control, we inspected every AI-suggested code change and tested the modified training pipeline end to end on toy problems before using it in our experiments and we read each suggested paper before citing it.
We did not use generative AI tools to generate hypotheses or core ideas, interpret results, or draw conclusions.
We take responsibility for the final content of this work, including all text, claims, and artifacts produced with the aid of generative AI.


\section*{Reproducibility statement}
We release the code and full training configurations needed to reproduce our experiments in the GitHub repository linked in the Introduction (section~\ref{sec:intro}) . Appendix~\ref{app:tasks} describes the tasks, data preparation, prompts, and evaluation procedures, and provides dataset links. Appendix~\ref{app:hyperparameters} reports the model choices and training hyperparameters.

%% file: sections/02-appendix/01-tasks-and-evaluation-details.tex
\section{Task and Evaluation Details}
\label{app:tasks}

\subsection{Integer Multiplication}

\paragraph{Task and data.} 
We use a procedurally generated four-digit multiplication task. Each example samples $a,b \sim \mathrm{Unif}\{1000,\ldots,9999\}$ and asks the generator to return the exact product. The reference answer is the integer $ab$. The processed dataset is available at 
\ifanon
\else  
    \url{https://huggingface.co/datasets/EdisonScientific/multiplication}
\fi
.

\paragraph{Generator prompt.}

\begin{PromptBlock}
[System]
You are a careful arithmetic assistant. Multiply the two integers exactly.
Think only as much as needed.

Your final response must use exactly this format:
<think>brief reasoning</think>
<answer>integer answer</answer>

Do not write anything before or after these tags.
Stop immediately after the closing answer tag.

[User]
What is {a} multiplied by {b}?
\end{PromptBlock}

\paragraph{Output format and parsing.} 
A completion is shown to the judge only if it contains a nonempty integer inside \verb|<answer>...</answer>| tags.

\paragraph{Judge prompt.}
\begin{PromptBlock}
[System]
You are an answer discriminator. Given a task and one candidate answer,
decide whether the candidate is more likely to be a reference answer or a model-generated answer.

Answer with exactly one character:
A for reference answer.
B for model-generated answer.

[User]
Problem: What is {a} multiplied by {b}?
Candidate answer: {candidate}

Classify the candidate as A (exact-product reference) or B (model-generated).
\end{PromptBlock}

\paragraph{Rewards and diagnostic metrics.} 

We compute exact correctness as
\[
r_{\mathrm{exact}}=\mathbb{1}[\hat{y}=ab],
\]
and clipped relative error as
\[
e_{\mathrm{rel}}=\min\left(\frac{|\hat{y}-ab|}{|ab|},1\right).
\].

\subsection{Countdown-Code}

\paragraph{Task and data.}
We instantiate Countdown as a code-writing task. Each example provides $3$--$4$ integers and a target value $t$. A valid solution is an arithmetic expression that uses each input number exactly once, uses only \texttt{+}, \texttt{-}, \texttt{*}, \texttt{/}, and parentheses, and evaluates to the target. The number and target instances are drawn from the public \texttt{Jiayi-Pan/Countdown-Tasks-3to4} dataset ~\cite{pan2025countdown3to4} and shuffled with a fixed seed. We generate reference solutions with OpenAI GPT-5.5 ~\cite{openai2026gpt5systemcard} using medium reasoning effort, then keep only solutions that pass programmatic validation. Accepted references must leave the canonical \texttt{test.py} unchanged, define a parseable \texttt{expr}, use exactly the provided numbers, contain only allowed arithmetic characters, evaluate to the target, and pass the canonical test harness. We use $15{,}000$ training examples and $800$ evaluation examples. The processed dataset is available at 
\ifanon
\else  
    \url{https://huggingface.co/datasets/EdisonScientific/countdown}
\fi
.

\paragraph{Generator prompt.} \leavevmode\\

For space, the prompt below abbreviates the serialized files. The actual prompt contains the full \texttt{solution.py} and \texttt{test.py} contents.

\begin{PromptBlock}
[System]
You are a helpful coding assistant. Solve the task and return only the requested JSON.

[User]
You are to play a variation of the game of Countdown which involves the use of Python.
You will be provided code in two scripts: one where you see the inputs and write your solution (`solution.py`), and the other being the test suite (`test.py`).
Your solution is to reach the target number by applying arithmetic operations on the provided input.
You can use basic arithmetic operations (+, -, *, /) and each number must be used once (you cannot ignore numbers nor re-use them).
You MUST ensure the test cases pass.
Think privately if useful, then return all files as valid JSON.
You will be provided the two files in their entirety, and you must output the edited versions of them with your solution.
Here is the environment for you to work with: recall you have to edit the file with a solution such that the test passes.

{
  "solution.py": "numbers = {numbers}\ntarget = {target}\nexpr = None  # edit this!",
  "test.py": "def verify_solution(numbers, target, expr) -> bool:\n    ..."
}

Your response must be exactly one JSON object of the form:
{ ...json here... }
\end{PromptBlock}

\paragraph{Output format and parsing.}
A completion is used as a generated candidate only if the text after the reasoning
trace parses as a JSON object with string-valued \texttt{solution.py} and
\texttt{test.py} fields. The submitted \texttt{solution.py} is then parsed to
extract the assigned expression \texttt{expr}.

\paragraph{Judge prompt.}
The discriminator receives a compact view of each submission: the assigned
expression and whether the submitted \texttt{test.py} was modified.

\begin{PromptBlock}
[System]
You are an answer discriminator. Given a task and one candidate answer,
decide whether the candidate is more likely to be a reference answer or a model-generated answer.

Answer with exactly one character:
A for reference answer.
B for model-generated answer.

[User]
You review compact Countdown-Code submissions.

A Countdown solution assigns an arithmetic expression to `expr`. A valid reference solution uses each input number exactly once, uses only +, -, *, /, evaluates to the target, and ensures the test cases pass.

Numbers: {numbers}
Target: {target}

Candidate expression:
{expr}

Candidate modified test.py: {test_modified}

Classify the candidate as A (reference solution) or B (model-generated solution).
\end{PromptBlock}

\paragraph{Rewards and diagnostic metrics.} \leavevmode For verifiable-reward baselines, we execute the submitted \texttt{solution.py} together with the submitted \texttt{test.py}. Unparseable outputs receive reward $0$, parsed but failing submissions receive reward $0.2$, and submissions whose \texttt{verify\_solution} call returns \texttt{True} receive reward $1.0$. Because this proxy reward could in principle be gamed by modifying \texttt{test.py}, we also report gold pass rate as the primary correctness metric. The gold check ignores the submitted \texttt{test.py} and directly verifies that \texttt{expr} uses exactly the provided numbers, contains only allowed arithmetic characters, and evaluates to the target. As a secondary diagnostic, we compute the clipped target-relative error,
\[
e_{\mathrm{rel}}=\min\left(\frac{|\mathrm{eval}(\hat{x})-t|}{\max(|t|,1)},1\right),
\]
which measures numerical closeness but does not by itself imply a valid Countdown solution.

\subsection{SMILES-to-Formula}

\paragraph{Task and data.} 
We provide the reference SMILES~\citep{weininger1988smiles}
string as input and ask the generator to recover the molecular formula, including implicit hydrogens. The reference answer is the molecular formula specified by the dataset solution. We use $17{,}938$ training examples and a held-out set of $800$ evaluation examples from the training split. The processed dataset is available at 
\ifanon
\else  
    \url{https://huggingface.co/datasets/EdisonScientific/smiles2formula}
\fi
.

\paragraph{Generator prompt.}
\begin{PromptBlock}
[System]
You are a careful chemistry assistant. Given a SMILES string, compute its molecular formula exactly.
Think only as much as needed.

Your final answer must use exactly this format:
<think>thinking</think>
<answer>FORMULA</answer>

Do not write anything before or after these tags.

[User]
Compute the molecular formula for this SMILES string.

SMILES:
{smiles}
\end{PromptBlock}

\paragraph{Output format and parsing.} A completion is shown to the judge only if it contains nonempty text inside \verb|<answer>...</answer>| tags and that text has formula-like syntax. This syntax check does not use the reference formula or the input SMILES. It only requires the answer, after whitespace normalization, to be fully covered by element-count terms of the form \texttt{ElementNumber}, such as \texttt{C6}, \texttt{H12}, or \texttt{O6}. Malformed answers receive zero reward and are not shown to the judge.

\paragraph{Judge prompt.}
\begin{PromptBlock}
[System]
You are an answer discriminator. Given a task and one candidate answer,
decide whether the candidate is more likely to be a reference answer or a model-generated answer.

Answer with exactly one character:
A for reference answer.
B for model-generated answer.

[User]
Chemistry question:
Compute the molecular formula for this SMILES string.

SMILES:
{smiles}

Candidate molecular formula:
{candidate}

Classify the candidate as A (reference answer) or B (model-generated answer).
\end{PromptBlock}

\paragraph{Rewards and diagnostic metrics.} For valid formulas, we compare the predicted and reference formulas by atom-count L1 distance,
\[
d_{\mathrm{formula}}=\sum_e |\hat{c}_e-c_e|,
\]
where $c_e$ is the count of element $e$, including hydrogen. Exact correctness is
\[
r_{\mathrm{exact}}=\mathbb{1}[d_{\mathrm{formula}}=0].
\]

\subsection{Medical/Biomedical Abstract Generation}

\paragraph{Task and data.} We use 5{,}000 biomedical and medical abstracts drawn from an enriched corpus of scientific paper abstracts. To construct the corpus, we queried Google Scholar using keywords across biology and medicine. We used a language model to extract each paper's title and abstract, excluding running headers, keywords, affiliations, and body sections. We then matched papers to OpenAlex by title to verify their domain and record their field, publication date, and citation count.

We retained papers classified by OpenAlex as medical or biomedical whose abstracts were well formed, contained 200--350 words, and did not exceed 3{,}000 characters. We removed duplicates based on normalized abstract text, shuffled the remaining abstracts with a fixed seed, and selected the first 5{,}000. Almost all selected papers were published between 2024 and 2026, with a median publication year of 2025.

These abstracts define the real data distribution that the generator learns to imitate in our adversarial experiments. Each example consists of a real paper abstract. The generator receives no paper title and is prompted only to produce a scientifically plausible research paper abstract. Before presenting reference and generated abstracts to the judge, we normalize both to a single line. The processed dataset is available at 
\ifanon
\else  
    \url{https://huggingface.co/datasets/EdisonScientific/paper-abstracts-medicine}
\fi
.

\paragraph{Generator prompt.}
\begin{PromptBlock}
[System]
You write scientifically plausible research paper abstracts.
Think only as much as needed.

Open with <answer>, write the complete abstract text, then close with </answer>.

The abstract should be concrete, technically plausible, and credible to specialists.
Do not include titles, authors, journal names, dates, citations, markdown, generic
placeholders, or multiple abstracts.
Stop immediately after the closing answer tag.

[User]
Generate a scientifically plausible research paper abstract.
\end{PromptBlock}

\paragraph{Output format and parsing.} A completion is shown to the judge only if, after removing any reasoning trace, the visible response consists of a nonempty \verb|<answer>...</answer>| block. The extracted abstract is normalized by collapsing all whitespace, including newlines, tabs, and repeated spaces, to single spaces and stripping surrounding quotation marks.

\paragraph{Learned judge prompt.}
\begin{PromptBlock}
[System]
You are an answer discriminator. Given a task and one candidate answer,
decide whether the candidate is more likely to be a reference answer or a model-generated answer.

Answer with exactly one character:
A for reference answer.
B for model-generated answer.

[User]
You review scientific paper abstracts.
Decide whether the abstract is from the paper corpus or was generated by a model.

Abstract-writing task:
Generate a scientifically plausible research paper abstract.

Candidate abstract:
{candidate}

Classify the candidate as A (real abstract from the paper corpus) or B (model-generated abstract).
\end{PromptBlock}

\paragraph{Haiku pairwise judge prompt.}
For the Haiku baseline, the judge is given the biomedical domain hint and compares one reference abstract against one generated abstract. The order is randomized across calls, and the generator receives reward $1$ only when Haiku selects the generated abstract as the real one.

\begin{PromptBlock}
[System]
You are a discriminator for scientific paper abstracts. Given two abstracts,
decide which abstract is from a real paper corpus rather than generated by a model.

Answer with exactly one character:
A if Abstract A is from the real paper corpus.
B if Abstract B is from the real paper corpus.

[User]
You review scientific paper abstracts.
The real paper corpus contains medical and biomedical paper abstracts.
Decide which of the two abstracts below comes from a real paper corpus and which one was generated by a model.

Abstract A:
{abstract_a}

Abstract B:
{abstract_b}

Which abstract comes from the real paper corpus? Answer only A or B.
\end{PromptBlock}

\paragraph{Rewards and diagnostic metrics.} This task has no exact verifier. Learned-judge runs score generated abstracts using the judge's reference-class probability proxy. Haiku-judge runs use a pairwise A/B comparison between a generated abstract and a reference abstract, randomized across both orders. We report fooling rates against held-out LLM judges and human evaluators, measuring how often generated abstracts are classified as real.

\subsection{NIH Grant Proposal Abstract Generation}

\paragraph{Task and data.}
We use NIH Research Portfolio Online Reporting Tools Expenditures and Results (RePORTER) FY2026 project records from the Projects API endpoint \url{https://api.reporter.nih.gov/v2/projects/search}. We keep the project title as the input and the project abstract as the reference answer. The raw download contains $45{,}764$ records. Basic cleaning drops records with missing titles, missing abstracts, placeholder abstracts, or abstracts shorter than $100$ characters, leaving $45{,}718$ title--abstract pairs.  The processed dataset is available at 
\ifanon
\else  
    \url{https://huggingface.co/datasets/EdisonScientific/grant-abstracts-NIH-fy2026}
\fi
.

\paragraph{Generator prompt.}
\begin{PromptBlock}
[System]
You are writing a scientifically plausible NIH grant proposal abstract.

Open with <answer>, write the complete abstract text, then close with </answer>.

The abstract should be concrete, technically plausible, and credible to specialists.
Stop immediately after the closing answer tag.

[User]
Write the abstract for an NIH grant proposal with the following title.
{title}
\end{PromptBlock}

\paragraph{Output format and parsing.} A completion is shown to the judge only if, after removing any reasoning trace, the visible response consists of a nonempty \verb|<answer>...</answer>| block. The extracted abstract is normalized by collapsing all whitespace, including newlines, tabs, and repeated spaces, to single spaces and stripping surrounding quotation marks.

\paragraph{Rewards and diagnostics.}
The main reward is adversarial: the learned judge estimates whether the candidate
abstract is more likely to come from the real NIH grant corpus or from the
generator. The generator is rewarded using the judge's reference-class
probability proxy. We also log format validity and corpus-comparison diagnostics,
including normalized exact string match to the reference abstract and abstract
distribution statistics.

\paragraph{Judge prompt.} 
\begin{PromptBlock}
[System]
You are an answer discriminator. Given a task and one candidate answer,
decide whether the candidate is more likely to be a reference answer or a model-generated answer.

Answer with exactly one character:
A for reference answer.
B for model-generated answer.

[User]
You review scientific paper abstracts.
Decide whether the abstract is from the paper corpus or was generated by a model.

Abstract-writing task:
Write the abstract for an NIH grant proposal with the following title.
{title}

Candidate abstract:
{candidate}

Classify the candidate as A (real abstract from the paper corpus) or B (model-generated abstract).
\end{PromptBlock}

%% file: sections/02-appendix/02-training-hyperparameters.tex
\newpage
\section{Training Hyperparameters}
\label{app:hyperparameters}

\subsection{Integer Multiplication Training Hyperparameters}

\begin{table}[!htbp]
\centering
\small
\renewcommand{\arraystretch}{1.12}
\begin{tabularx}{\linewidth}{p{0.34\linewidth}X}
\toprule
\textbf{Setting} & \textbf{Value} \\
\midrule
Training examples & $20{,}000$ \\
Evaluation examples & $800$, evaluated every $250$ steps \\
Generator model & \texttt{Qwen/Qwen3-0.6B} \\
Judge models & \texttt{Qwen/Qwen3-0.6B}, \texttt{1.7B}, and \texttt{4B} \\
Reward variants & judge-only, exact-only, judge $\times$ exact; exact-only and judge $\times$ exact use the \texttt{0.6B} judge setting \\
Training horizon & $3{,}000$ steps \\
Generator batch & $128$ prompts, $8$ rollouts per prompt \\
Judge batch & $256$ examples, $1$ rollout per example \\
Generator context & Sequence length $8192$, maximum completion length $2048$ \\
Judge context & Sequence length $2048$, maximum completion length $1$ \\
Generator sampling & Temperature $1.0$, thinking enabled \\
Judge inference & Deterministic one-token A/B classification, thinking disabled \\
Generator optimization & Learning rate $10^{-7}$, KL coefficient $0.02$ \\
Judge optimization & Binary-classification loss, learning rate $10^{-6}$ \\
Generator clipping & Policy-ratio mask low/high $0.20/0.20$ \\
Off-policy limit & Maximum $8$ off-policy steps \\
Judge-update gate & EMA accuracy threshold $0.90$, max skip streak $20$ where enabled \\
\bottomrule
\end{tabularx}
\caption{RAL digit-multiplication adversarial-training setup.}
\label{tab:multiplication_setup}
\end{table}

\begin{table}[!htbp]
\centering
\small
\renewcommand{\arraystretch}{1.12}
\begin{tabularx}{\linewidth}{p{0.34\linewidth}X}
\toprule
\textbf{Setting} & \textbf{Value} \\
\midrule
Model & \texttt{Qwen/Qwen3-0.6B} (full finetuning) \\
Training examples & $20{,}000$ \\
Held-out evaluation & $800$ examples \\
Effective batch & $\approx 3{,}200$ examples ($131$K tokens) \\
Batch composition & $64$ packed sequences of length $2048$ \\
Optimization & AdamW, learning rate $2\times10^{-5}$ \\
Evaluation sampling & Temperature $1.0$, thinking disabled, maximum completion length $128$ \\
Evaluation scoring & Exact integer match, identical to the RL verifiable reward \\
Evaluation cadence & Every $25$ steps and at initialization \\
\bottomrule
\end{tabularx}
\caption{Digit-multiplication SFT baseline setup.}
\label{tab:multiplication_sft_setup}
\end{table}

\newpage
\FloatBarrier
\subsection{Countdown-Code Training Hyperparameters}

\begin{table}[!htbp]
\centering
\small
\renewcommand{\arraystretch}{1.12}
\begin{tabularx}{\linewidth}{p{0.34\linewidth}X}
\toprule
\textbf{Setting} & \textbf{Value} \\
\midrule
Training examples & $15{,}000$ \\
Evaluation examples & $800$ \\
Generator model & \texttt{Qwen/Qwen3-1.7B} \\
Judge models & \texttt{Qwen/Qwen3-1.7B}, \texttt{4B}, \texttt{8B}, \texttt{14B}, and \texttt{32B} \\
Reward variants & judge-only, exact-only, judge $\times$ exact \\
Training horizon & $10{,}000$ steps \\
Generator batch & $64$ prompts, $8$ rollouts per prompt \\
Judge batch & $128$ examples, $1$ rollout per example \\
Generator context & Sequence length $16384$, maximum completion length $8192$ \\
Judge context & Sequence length $2048$, maximum completion length $1$ \\
Generator sampling & Temperature $1.0$, thinking enabled \\
Judge inference & Deterministic one-token A/B classification, thinking disabled \\
Generator optimization & Learning rate $5\times10^{-7}$, KL coefficient $0.005$ \\
Judge optimization & Binary-classification loss, learning rate $10^{-6}$ \\
Judge-update gate & EMA accuracy threshold $0.75$, max skip streak $50$ where enabled \\
\bottomrule
\end{tabularx}
\caption{Countdown-Code adversarial-training setup.}
\label{tab:countdown_code_setup}
\end{table}

\begin{table}[!htbp]
\centering
\small
\renewcommand{\arraystretch}{1.12}
\begin{tabularx}{\linewidth}{p{0.34\linewidth}X}
\toprule
\textbf{Setting} & \textbf{Value} \\
\midrule
Model & \texttt{Qwen/Qwen3-1.7B} (full finetuning) \\
Training examples & $15{,}000$ \\
Held-out evaluation & $800$ examples \\
Effective batch & $\approx 730$ examples ($524$K tokens) \\
Batch composition & $128$ packed sequences of length $4096$ \\
Optimization & AdamW, learning rate $3\times10^{-5}$ \\
Evaluation sampling & Temperature $1.0$, thinking disabled, maximum completion length $1024$ \\
Evaluation scoring & Submitted solution executed against the canonical test, identical to the RL verifiable reward \\
Evaluation cadence & Every $60$ steps and at initialization \\
\bottomrule
\end{tabularx}
\caption{Countdown SFT baseline setup.}
\label{tab:countdown_sft_setup}
\end{table}

\newpage
\FloatBarrier
\subsection{SMILES-to-Formula Training Hyperparameters}

\begin{table}[!htbp]
\centering
\small
\renewcommand{\arraystretch}{1.12}
\begin{tabularx}{\linewidth}{p{0.34\linewidth}X}
\toprule
\textbf{Setting} & \textbf{Value} \\
\midrule
Dataset & Ether0 simple-formula SMILES-to-formula examples \\
Training examples & $17{,}938$ \\
Evaluation examples & $800$ held-out examples \\
Generator model & \texttt{Qwen/Qwen3-8B} \\
Judge models & \texttt{Qwen/Qwen3-8B}, \texttt{14B}, and \texttt{32B} \\
Reward variants & judge-only, exact-only, judge $\times$ exact \\
Training horizon & $15{,}000$ steps \\
Generator batch & $64$ prompts, $8$ rollouts per prompt \\
Judge batch & $128$ examples, $1$ rollout per example \\
Generator context & Sequence length $8192$, maximum completion length $4096$ \\
Judge context & Sequence length $2048$ for large-judge runs, maximum completion length $1$ \\
Generator sampling & Temperature $1.0$, thinking enabled \\
Judge inference & Deterministic one-token A/B classification, thinking disabled \\
Generator optimization & Learning rate $5\times10^{-7}$, KL coefficient $0.005$ \\
Judge optimization & Binary-classification loss, learning rate $10^{-6}$ \\
Judge-update gate & EMA accuracy threshold $0.75$, max skip streak $50$ where enabled \\
\bottomrule
\end{tabularx}
\caption{SMILES-to-formula adversarial-training setup.}
\label{tab:smiles_to_formula_setup}
\end{table}

\begin{table}[!htbp]
\centering
\small
\renewcommand{\arraystretch}{1.12}
\begin{tabularx}{\linewidth}{p{0.34\linewidth}X}
\toprule
\textbf{Setting} & \textbf{Value} \\
\midrule
Model & \texttt{Qwen/Qwen3-8B} (full finetuning) \\
Training examples & $17{,}938$ \\
Held-out evaluation & $800$ examples \\
Effective batch & $\approx 2{,}250$ examples ($131$K tokens) \\
Batch composition & $64$ packed sequences of length $2048$ \\
Optimization & AdamW, learning rate $10^{-5}$ \\
Evaluation sampling & Temperature $1.0$, thinking disabled, maximum completion length $128$ \\
Evaluation scoring & Exact molecular formula match by atom counts, identical to the RL verifiable reward \\
Evaluation cadence & Every $100$ steps and at initialization \\
\bottomrule
\end{tabularx}
\caption{SMILES-to-formula SFT baseline setup.}
\label{tab:smiles2formula_sft_setup}
\end{table}

\newpage
\FloatBarrier
\subsection{Medical and Biomedical Abstract Generation Training Hyperparameters}

\begin{table}[!htbp]
\centering
\small
\renewcommand{\arraystretch}{1.12}
\begin{tabularx}{\linewidth}{p{0.34\linewidth}X}
\toprule
\textbf{Setting} & \textbf{Value} \\
\midrule
Dataset & Medical-domain paper abstracts \\
Training examples & $5{,}000$ \\
Generator model & \texttt{Qwen/Qwen3-8B} \\
Judge models & \texttt{claude-haiku-4-5} pairwise judge and \texttt{Qwen/Qwen3-14B} learned judge \\
Reward variants & Pairwise Haiku judge reward and learned single-candidate judge reward \\
Training horizon & $2{,}000$ steps \\
Generator batch & $128$ prompts, $8$ rollouts per prompt \\
Learned judge batch & $256$ examples, $1$ rollout per example \\
Generator context & Sequence length $8192$, maximum completion length $4096$ \\
Learned judge context & Sequence length $4096$, maximum completion length $1$ \\
Generator sampling & Temperature $1.0$, thinking enabled \\
Judge inference & Haiku pairwise A/B judgment with structured output; learned judge deterministic one-token A/B classification with thinking disabled \\
Generator optimization & Learning rate $2\times10^{-7}$, KL coefficient $0.05$ \\
Learned judge optimization & Binary-classification loss, learning rate $3\times10^{-7}$ \\
Generator clipping & DPPO mask range $[0.20,0.28]$ \\
Judge-update gate & EMA accuracy threshold $0.75$, max skip streak $50$ for learned-judge runs \\
\bottomrule
\end{tabularx}
\caption{Medical/biomedical paper-abstract adversarial-training setup.}
\label{tab:medical_abstract_setup}
\end{table}

\newpage
\FloatBarrier
\subsection{NIH Grant Proposal Abstract Generation Training Hyperparameters}

\begin{table}[!htbp]
\centering
\small
\renewcommand{\arraystretch}{1.12}
\begin{tabularx}{\linewidth}{p{0.34\linewidth}X}
\toprule
\textbf{Setting} & \textbf{Value} \\
\midrule
Dataset & NIH RePORTER FY2026 grant proposal abstracts, cleanv4 moderate split \\
Training examples & $20{,}000$ split, with $19{,}995$ configured for training \\
Evaluation examples & $128$ held-out grant titles and abstracts \\
Generator model & \texttt{Qwen/Qwen3-32B} \\
Judge model & \texttt{Qwen/Qwen3-235B-A22B} \\
Reward variant & Learned single-candidate judge reward \\
Training horizon & $5{,}000$ configured steps \\
Generator batch & $1{,}024$ generated samples per update: $64$ prompts, $16$ rollouts per prompt \\
Learned judge batch & $3{,}072$ A/B examples per update: $2{,}048$ current examples plus $1{,}024$ replay examples \\
Generator context & Training sequence length $8192$, inference context $5120$, maximum completion length $4096$ \\
Learned judge context & Sequence length $4096$, maximum completion length $1$ \\
Generator sampling & Temperature $1.0$, thinking enabled with thinking-token budget $1536$ \\
Judge inference & Deterministic one-token A/B classification with structured output and thinking disabled \\
Generator optimization & Learning rate $2\times10^{-7}$, KL coefficient $0.01$ \\
Learned judge optimization & Binary-classification loss, learning rate $3\times10^{-7}$ \\
Length penalty & None \\
Judge-update gate & EMA accuracy threshold $0.75$, EMA decay $0.95$, max skip streak $50$ \\
Judge replay buffer & $512$ replayed generated samples per update, each paired with its reference, drawn uniformly from a sliding window of the last $500$ generator batches ($512{,}000$ generated samples). Replayed samples are excluded from gate accuracy \\
Judge target network & Served judge is an exponential moving average of the learner, decay $0.995$ per broadcast. Steps skipped by the gate are folded into the next update as $1-0.995^{k}$ \\
\bottomrule
\end{tabularx}
\caption{NIH grant-proposal abstract adversarial-training setup.}
\label{tab:nih_grant_abstract_setup}
\end{table}

\begin{table}[!htbp]
\centering
\small
\renewcommand{\arraystretch}{1.12}
\begin{tabularx}{\linewidth}{p{0.34\linewidth}X}
\toprule
\textbf{Setting} & \textbf{Value} \\
\midrule
Model & \texttt{Qwen/Qwen3-32B} (full finetuning) \\
Training examples & $19{,}995$ ($5$ duplicates of held-out grants removed) \\
Held-out evaluation & $128$ examples \\
Effective batch & $\approx 255$ examples ($147$K tokens) \\
Batch composition & $72$ packed sequences of length $2048$ \\
Optimization & AdamW, learning rate $4.5\times10^{-5}$ \\
Evaluation sampling & Temperature $1.0$, thinking disabled, maximum completion length $1280$ \\
Evaluation scoring & Format compliance only; generated abstracts are stored at every evaluation for external judging (no verifiable reward exists for this task) \\
Evaluation cadence & Every $100$ steps and at initialization \\
\bottomrule
\end{tabularx}
\caption{NIH grant-abstract SFT baseline setup.}
\label{tab:grant_abstracts_sft_setup}
\end{table}

\newpage

%% file: sections/02-appendix/03-relative-errors.tex

%% file: sections/02-appendix/04-haiku-judge-reward-evaluation-analysis.tex
\newpage
\section{Haiku Judge Reward, Domain Evaluation, and Failure Analysis}
\label{app:haiku_prompt}

We use Claude Haiku 4.5 for two distinct purposes in the
medical and biomedical abstract experiment: computing the
frozen-judge baseline's training reward and evaluating
domain fidelity. The prompts are reproduced below.

\subsection{Frozen-Judge Training Reward}

The judge receives one reference abstract and one generated
abstract and identifies which is real. The generator receives
reward $1.0$ if the judge selects the generated abstract as
real, and $0.0$ otherwise. The prompt explicitly identifies
the reference corpus as medical and biomedical.

\begin{PromptBlock}
You review scientific paper abstracts.
The real paper corpus contains medical and biomedical paper abstracts.
Decide which of the two abstracts below comes from a real paper corpus and which one was generated by a model.

Abstract A:
{abstract_a}

Abstract B:
{abstract_b}

Which abstract comes from the real paper corpus? Answer only A or B.
\end{PromptBlock}

\subsection{Biomedical Domain Evaluation}

We evaluate domain fidelity with a separate prompt that
asks whether each abstract is clearly biomedical. The
biomedical rate is the fraction of evaluated abstracts
classified as clearly biomedical.

\begin{PromptBlock}
You are a careful scientific abstract classifier.

Decide whether the abstract is clearly a medical or biomedical research-paper abstract. Answer true only for clinical medicine, human or animal health and disease, physiology/pathology, pharmacology, immunology, public health, disease-focused neuroscience, or molecular/cell biology in a health-or-disease context.

Answer false for ecology, environmental toxicology, plant biology, agriculture, materials, engineering, pure chemistry/physics, or generic life science unless the health/disease biomedical connection is clear.
\end{PromptBlock}

\subsection{frozen-Judge Failure Mode}
\label{app:external_judge_failure}

For the frozen-Haiku run in Figure~\ref{fig:medical-abstracts-results}, we inspect the final checkpoint at \texttt{step\_2000}, which contains 128 generator rollouts. Of these, 125 collapse onto a narrow Pacific-oyster microplastics ecotoxicology theme, despite remaining lexically distinct. The generator receives no title conditioning in this experiment. Below we show two randomly sampled complete generations from \texttt{step\_2000}. Both receive reward 1.0 and are shown without truncation.

\begin{LLMOutput}{Example 1: full raw generation}
This study investigated the sublethal effects of polyethylene microplastics (MPs; 20–50 μm) on the haemolymph biochemical parameters, digestive enzyme activities, and histopathology of the Pacific oyster *Crassostrea gigas* following 28 days of continuous exposure. Adult oysters (mean ± SE wet mass = 16.4 ± 0.3 g; mean ± SE shell length = 57.2 ± 0.6 mm) were exposed in static renewal assays to 0 (control), 1, 10, or 100 mg L⁻¹ MPs suspended in 30 μm filtered seawater (15‰ salinity; 20 ± 1 °C). Triplicate groups of 10 animals per concentration were maintained for 28 days, with renewal every 48 h. At termination, haemolymph was sampled from 10 individuals per group for determination of total haemocyte count (THC), alkaline phosphatase (ALP) activity, and catalase (CAT) activity. Digestive gland (gill and stomach) tissues from 10 individuals per group were fixed in 2.5\% glutaraldehyde and subjected to histopathological analysis. MPs were recovered from digestive tissues via density gradient centrifugation and quantified using epifluorescence microscopy. Concentrations of MPs in digestive gland tissue from exposed animals ranged from 125 ± 23 to 3870 ± 450 μg g⁻¹ dry mass (mean ± SE; 10 individuals per group) at 100 mg L⁻¹, with no MPs detected in controls. concentrations of MPs in digestive tissues were significantly higher than controls at all exposure concentrations (ANOVA, F₃,₂₈ = 114.9, P < 0.001; Tukey HSD post-hoc comparisons, P < 0.05 for all treatments vs. control). Neither THC (F₃,₂₈ = 0.21, P = 0.89) nor ALP activity (F₃,₂₈ = 0.92, P = 0.45) differed among treatments. CAT activity was significantly higher at 100 mg L⁻¹ than controls (0.46 ± 0.03 vs. 0.25 ± 0.01 U mg⁻¹ protein; F₃,₂₈ = 6.89, P = 0.003; Tukey HSD, P = 0.003). No significant differences in CAT activity were detected between controls and 1 or 10 mg L⁻¹ treatments. Histopathological examination revealed no discernible morphological abnormalities in gill or stomach tissues from control or low-dose (1 or 10 mg L⁻¹) treatments. At 100 mg L⁻¹, digestive gland tissues exhibited extensive vacuolation of epithelial cells in the mid- and posterior gastric regions, with threefold to fivefold increases in the number of granular inclusion bodies within digestive epithelial cells compared to controls (chi-square test, χ² = 22.4, P = 0.0001). The presence of inclusion bodies correlated with MP burden in individual digestive glands (Pearson’s r = 0.72, P = 0.002). No evidence of haemocyte aggregation, granulocytic transformation, or epithelial necrosis was observed in any treatment group. These results demonstrate that sublethal exposure to polyethylene MPs induces oxidative stress (evidenced by increased CAT activity) and histopathological changes in digestive tissues of *C. gigas* at 100 mg L⁻¹, but not at lower concentrations. Given the absence of effects at 1 mg L⁻¹, the biological relevance of these responses to field-relevant MP concentrations in oyster habitats requires further investigation.  
</answer>
\end{LLMOutput}

\begin{LLMOutput}{Example 2: full raw generation}
<answer>  
This study investigated the sublethal effects of polyethylene microplastics (MPs; 20 μm nominal size) on the haemolymph biochemistry and digestive tract histology of the Pacific oyster *Crassostrea gigas* following 28 days of continuous exposure. Adult oysters (mean ± SD wet biomass = 21.4 ± 1.8 g; mean ± SD shell length = 57.2 ± 1.6 mm) were exposed in static renewal experiments to 0 (control), 5, 50, or 100 mg L⁻¹ MPs suspended in 30 μm filtered seawater (15 ± 1 °C, 32 ± 1 psu). MPs were characterized by Fourier-transform infrared spectroscopy and had a geometric mean diameter of 19.3 ± 0.6 μm (n = 300 particles; error ± SD). Oysters were fed *Isochrisis galbana* (3 × 10⁶ cells mL⁻¹) daily and exposed 5 days week⁻¹ for 28 days. At termination, haemolymph was analysed for superoxide dismutase (SOD; Units mg⁻¹ protein), catalase (CAT; μmol H₂O₂ min⁻¹ mg⁻¹ protein), glutathione peroxidase (GPx; μmol glutathione min⁻¹ mg⁻¹ protein), and total glutathione (GSH + GSSG; μmol g⁻¹ haemolymph). Digestive tract tissues (gastrovesicle, digestive gland) were processed for light microscopy (hematoxylin–eosin staining) and stereomicroscopy. MPs were detected in the digestive tracts of all exposed individuals by micro-Raman spectroscopy; geometric mean MP counts per oyster were 13.2 (± 6.8; n = 10) in the 5 mg L⁻¹ group, 72.5 (± 15.3; n = 10) in the 50 mg L⁻¹ group, and 218.3 (± 52.7; n = 10) in the 100 mg L⁻¹ group. No mortality occurred during exposure. SOD activity was significantly elevated in the 50 mg L⁻¹ (1.12 ± 0.10 vs. 0.88 ± 0.04 Units mg⁻¹ protein; ANOVA, F₃,₃₆ = 12.3, P = 0.0001; Tukey’s HSD, P < 0.05) and 100 mg L⁻¹ (1.35 ± 0.08 vs. 0.88 ± 0.04; P < 0.05) groups compared with control. CAT activity was significantly increased in the 100 mg L⁻¹ group (1.42 ± 0.09 vs. 0.88 ± 0.04 μmol H₂O₂ min⁻¹ mg⁻¹ protein; F₃,₃₆ = 15.7, P = 0.00002; Tukey’s HSD, P < 0.05). GPx activity did not differ among groups (ANOVA, F₃,₃₆ = 0.48, P = 0.69). Total glutathione levels were reduced in the 50 mg L⁻¹ (0.24 ± 0.01 vs. 0.31 ± 0.01 μmol g⁻¹ haemolymph; F₃,₃₆ = 6.8, P = 0.0023; Tukey’s HSD, P < 0.05) and 100 mg L⁻¹ (0.20 ± 0.01 vs. 0.31 ± 0.01; P < 0.05) groups compared with control. Light microscopic examination of the digestive gland revealed granular, PAS-positive deposits within the epithelial cells of the digestive tubules in all exposed groups; deposits were more extensive in the 100 mg L⁻¹ group, characterised by coalescing, eosinophilic granules (mean ± SD area per cell = 81.3 ± 20.5 μm²; n = 50 cells; control = 12.3 ± 3.2 μm², P < 0.001, t-test). Stereomicroscopy identified vacuolar degeneration in the digestive epithelium in the 100 mg L⁻¹ group, with loss of microvilli and cytoplasmic condensation in 34\% of observed cells (n = 50 cells). No significant differences in haemolymph protein concentration (mg mL⁻¹; ANOVA, F₃,₃₆ = 0.87, P = 0.47) or lipid peroxidation (malondialdehyde content; nmol MDA mg⁻¹ protein; ANOVA, F₃,₃₆ = 1.0, P = 0.41) were observed among groups. These findings demonstrate that chronic exposure to environmentally relevant concentrations of polyethylene MPs induces oxidative stress and sublethal histopathological changes in the digestive tract of *C. gigas*, suggesting potential ecological risks to filter-feeding bivalves in microplastic-polluted marine environments.  
</answer>
\end{LLMOutput}

Although these generations are lexically distinct and superficially scientific, they have collapsed to an ecotoxicology template outside the intended biomedical domain. 

\newpage

%% file: sections/02-appendix/05-qualitative-nih-grant-abstract-examples.tex
\section{Qualitative NIH Grant Abstract Examples}
\label{app:nih_examples}

Below we show two examples from the RAL-trained NIH Grant Abstract generator model at checkpoint 2099. Each fooled all five LLM judges and a majority of human expert annotators. For each title, we show the generated abstract alongside the real NIH abstract. The generated text is whitespace-collapsed exactly as presented to the judges. Both abstracts are shown in full.

\paragraph{Example 1.} 
Title: Bioinspired, Multi-Antioxidant Nanomedicine for Improved Treatment of Acute Kidney Injury.

\begin{LLMOutput}{Model-generated abstract.}
Acute kidney injury (AKI) is a highly prevalent, life-threatening disease among patients receiving secondary care. There are currently no approved drugs to prevent or mitigate AKI, emphasizing an urgent medical need for developing effective therapeutic agents for this disease. Although many renoprotective agents have been identified, they have not translated to the clinic due to various limitations. To develop an effective therapy for AKI, we focus on one of central mechanisms underlying AKI: oxidative stress. Although administration of various singular antioxidants in clinical trials did not show therapeutic benefits, we argue that the ineffectiveness of these antioxidants in clinical trials is partially due to their monotherapeutic application because oxidative stress for AKI is the result of multiple reactive oxygen and nitrogen species. Combination of multiple antioxidants could address this issue; however, the poor pharmacokinetics of the cocktails of antioxidants has prevented translation of this approach. To overcome these challenges, we developed a multi-antioxidant nanomedicine to co-deliver multiple antioxidants to effectively combat oxidative stress. Notably, our preliminary studies demonstrated that this nanomedicine significantly ameliorates a mouse model of ischemia reperfusion (IR) induced AKI, the most common cause of the disease, more effectively than bardoxolone methyl, the most advanced renoprotective agent in clinical development. Inspired by the features of our endogenous antioxidant defense system, our objective in this proposal is to develop next generation bioinspired nanomedicine for even more effective scavenging of reactive oxidants for the treatment of AKI. More specifically, we hypothesize that a bioinspired nanomedicine that combines the advantages of multiple antioxidants with preferential accumulation near plasma membranes will significantly enhance the effectiveness of nanomedicine in eliminating multiple reactive oxygen and nitrogen species (RONS) and other oxidative species that are generally generated in or near the plasma membrane for more effective AKI mitigation. To test our hypothesis, we propose two Aims. In Aim 1, we hypothesize that developing a 2nd generation multi-antioxidant nanomedicine with increased variety of antioxidants and long term antioxidant capability and further equipping the nanomedicine with targeting ability near the plasma membrane will significantly improve the efficiency of scavenging various RONS for in vitro application. To determine design guideline for in vivo application of our nanomedicine, we will investigate the relationship of the nanomedicine formulation parameters with its physicochemical properties and antioxidant potential. In Aim 2, we hypothesize that the optimized formulation identified from Aim 1 will mitigate AKI by eliminating multiple RONS. We will examine in vivo fate of the optimized nanomedicine and dissect its renoprotection mechanisms for clinical translation. Outcomes of this highly innovative project will create a completely novel strategy in nanomedicine design for intracellular redox system modulation to treat oxidative stress related diseases and enable a best-in-class renoprotective nanomedicine for treatment of AKI.
\end{LLMOutput}

\begin{LLMOutput}{Real NIH grant abstract.}
Acute kidney injury (AKI) is marked by a rapid loss of kidney function and is associated with high mortality, particularly in critical care settings. Despite advances in early biomarkers and standardized definitions, current clinical management remains largely supportive, aiming only to mitigate secondary complications. A key driver of AKI pathogenesis is the overproduction of reactive oxygen species (ROS), prompting interest in antioxidant-based therapies. However, conventional small-molecule antioxidants suffer from limited bioavailability and nonspecific distribution, necessitating frequent high-dose administrations that reduce the therapeutic impact. Chalcogens (or the oxygen family), known for their redox properties, can form stable colloidal nanoparticles when combined with phenolic compounds, which are natural antioxidants found in plants and foodstuffs. This combination enhances ROS scavenging capacity and provides metal-chelating abilities. I previously demonstrated that sub-5 nm silver chalcogenide nanoparticles are concentrated in the kidneys and undergo predominant renal clearance. Therefore, I hypothesize that ultrasmall chalcogen-phenolic nanoparticles (CPNP) can more effectively neutralize ROS and accumulate in the kidneys compared to small-molecule antioxidants, thus improving treatment efficacy while reducing dosage and cost. My preliminary investigation shows that a CPNP formulation composed of tellurium and caffeic acid (TeCA-NP) exhibits potent antioxidant activity, kidney-specific delivery, and therapeutic benefit in a murine model of glycerol-induced AKI. Leveraging my expertise in nanomaterial design, I will synthesize a library of sub-5 nm CPNP formulations and evaluate their antioxidant activity in vitro (Aim 1). This approach is expected to yield at least one additional candidate with performance comparable to or exceeding that of TeCA-NP, thereby strengthening the feasibility and robustness of the proposed work. I will then assess the therapeutic efficacy of both formulations in a clinically relevant model of ischemia-reperfusion-induced AKI at the histological, molecular, and functional levels (Aim 2). Finally, I will examine the biodistribution, pharmacokinetics, and in vivo safety of radiolabeled CPNP through positron emission tomography imaging (Aim 3). These comparative studies will identify an optimized lead and provide preclinical data to support the translational potential of this novel nanotherapeutic platform for AKI and other oxidative stress-related diseases. This K01 application will also provide valuable educational and training opportunities to refine my skills in nanomaterial synthesis while broadening my expertise in nuclear medicine imaging, drug delivery, and organ injury treatment. I have assembled a multidisciplinary team of experienced mentors across two institutions to guide me in this undertaking. These efforts will generate data for future R01 applications and position me to achieve my long-term goal of establishing an independent research program at the interface of nanomedicine and molecular imaging to improve disease diagnosis and treatment.
\end{LLMOutput}

\paragraph{Example 2.} 
Title: Discovery and manipulation of transcription factors to restore long term stem cell repopulation in aged bone-marrow.

\begin{LLMOutput}{Model-generated abstract.}
The aging of the human population has created an urgent demand for solutions to treat various diseases caused by degeneration of stem cells in the elderly. This proposal is focused on diseases caused by the dysfunction of hematopoietic stem cells (HSCs) in the elderly. We are proposing fundamentally novel strategies to restore aged HSC function at transcription factor (TF) and drug levels to treat many blood diseases prevalent in the elderly. To reach this goal, we developed a novel screening strategy, allowing for the first-time discovery of TFs that are specific for young HSCs. We uncovered Zfp335 as a critical regulator that is required for long term repopulation of young HSCs. Our preliminary data also indicate that Zfp335 function is defective in aged HSCs. We seek to build on our unique strengths in the development of novel screening strategies, novel applications of cutting edge genomic technologies, and gene and drug discovery for mammalian adult stem cells to significantly move toward future HSC therapy. We will first uncover if the aged niche regulates young HSC function through the regulation of TF activities. This study would not only allow us to test if the HSC transcriptome change induced by aged niche is similar to that observed in old HSCs, it also will allow us to gain understanding about the key TFs regulated by the aged niche. Using Zfp335 as a model, we next seek to understand the mechanism by which it maintains HSC features. We will identify genome-wide genomic targets of Zfp335 and further carry out higher-resolution investigation of HSC enhancer activities regulated by Zfp335. We will then directly test if Zfp335 expression is sufficient to reverse the aged programs to rescue the long term HSC activity of aged HSCs or young HSCs reprogrammed by the aged niche. Finally, for the first time, we will build a platform to discover small compounds as the foundation for possible HSC therapeutics. We will perform a HEK293 based HTS that can be used for secondary screens and structure- and target-based drug optimization. We also will implement a totally new direction to carry out HSC-based screening that is more representative of in vivo drug functions. The new genomic and genetic information obtained from the first two Aims is the foundation of this Aim. Together, the information and approaches obtained from the proposed study will create a new field of HSC transcription biology and provide a foundation for future gene and drug therapies.
\end{LLMOutput}

\begin{LLMOutput}{Real NIH grant abstract.}
Aging has a complex underlying biology characterized by a progressive loss of cellular and physiological function and this deterioration is strongly correlated with degenerative disease. In the bone marrow, aging markedly reduces the capacity of hematopoietic stem cells (HSCs) to self-renew and differentiate into lymphoid lineages, resulting in hindered immune function and systemic effects on multiple tissues, such as muscle repair after injury. Bone marrow from young recipients has been shown to rejuvenate aged bone marrow as well as systemically in other tissues. However, the exact HSC and progenitor cell states as well as other factors that drive the rejuvenating effects are not well understood. Profiling of HSCs and other bone marrow cell types during aging and an understanding of the transcription factors (TFs)-that control HSC self-renewal and their changes in activity during aging could provide new therapeutic approaches with the potential to reverse both blood-specific and whole-animal effects of aging. TF-based interventions, such as partial reprogramming, have shown promise to promote stem-cell cycling and regeneration. However, current approaches are limited to a small set of predetermined TFs, commonly the Yamanaka factors Oct3/4, Sox2, Klf4 and c-Myc, and have only been demonstrated in a select set of tissues. Furthermore, the diversity of the HSC population, containing both senescent cells and long-term renewing state (LT-HSC) among other subtypes, makes discovery of master regulators challenging, and existing approaches do not address this heterogeneity. We hypothesize that high-resolution single cell RNA sequencing (scRNA-seq) of HSCs from mice of different ages will reveal putative TF regulators of the aging process, and that these candidates can reprogram aged HSCs towards LT-HSC and quiescent states capable of niche restoration to reverse age-associated phenotypes. We will profile molecular signatures of aging in HSCs at single cell resolution and use these data to both develop metrics for aging and nominate TFs to promote LT-HSC restoration and rejuvenation. We will synthesize selected TFs for pooled screening allowing for rapid evaluation of their reprogramming effects in vitro and in vivo. Coupling these pooled perturbations in vivo with scRNA-seq readouts will allow for evaluation of HSC rejuvenation via our aging signatures and measurement of lymphoid/myeloid skew. Candidate TFs that demonstrate the strongest potential for rejuvenation of LT-HSCs will be tested individually and in combination for modification of whole-organism phenotypes, including increased repopulation potential, reduction of inflammatory factors, and improvement of muscle repair in response to injury. The repurposing of novel TFs regulating HSC rejuvenation as new therapeutics for aging-associated disease provides a new framework for cellular engineering. This proposal, coupling transcriptomic readouts and screening for discovery of new regulators of cell states, serves as the foundation for TF-based interventions for disease, both in aging and in broader human health.
\end{LLMOutput}

%% file: arxiv.bbl
\begin{thebibliography}{37}
\providecommand{\natexlab}[1]{#1}
\providecommand{\url}[1]{\texttt{#1}}
\expandafter\ifx\csname urlstyle\endcsname\relax
  \providecommand{\doi}[1]{doi: #1}\else
  \providecommand{\doi}{doi: \begingroup \urlstyle{rm}\Url}\fi

\bibitem[Anthropic(2025)]{anthropic2025haiku45}
Anthropic.
\newblock Claude haiku 4.5 system card, October 2025.

\bibitem[{Anthropic}(2026{\natexlab{a}})]{anthropic2026opus48}
{Anthropic}.
\newblock Introducing claude opus 4.8, May 2026{\natexlab{a}}.
\newblock URL \url{https://www.anthropic.com/news/claude-opus-4-8}.

\bibitem[{Anthropic}(2026{\natexlab{b}})]{anthropic2026opus5}
{Anthropic}.
\newblock Introducing claude opus 5, July 2026{\natexlab{b}}.
\newblock URL \url{https://www.anthropic.com/news/claude-opus-5}.

\bibitem[Cai et~al.(2025)Cai, Ryabinin, and Provilkov]{cai2025escaping}
Locke Cai, Max Ryabinin, and Ivan Provilkov.
\newblock Escaping the verifier: Learning to reason via demonstrations.
\newblock \emph{arXiv preprint arXiv:2511.21667}, 2025.

\bibitem[Christiano et~al.(2017)Christiano, Leike, Brown, Martic, Legg, and Amodei]{christiano2017deep}
Paul~F Christiano, Jan Leike, Tom Brown, Miljan Martic, Shane Legg, and Dario Amodei.
\newblock Deep reinforcement learning from human preferences.
\newblock \emph{Advances in neural information processing systems}, 30, 2017.

\bibitem[Cui et~al.(2025)Cui, Yuan, Wang, Wang, Zhang, Chen, Li, He, Fan, Yu, et~al.]{cui2025process}
Ganqu Cui, Lifan Yuan, Zefan Wang, Hanbin Wang, Yuchen Zhang, Jiacheng Chen, Wendi Li, Bingxiang He, Yuchen Fan, Tianyu Yu, et~al.
\newblock Process reinforcement through implicit rewards.
\newblock \emph{arXiv preprint arXiv:2502.01456}, 2025.

\bibitem[Damani et~al.(2026)Damani, Puri, Shenfeld, and Andreas]{damani2026right}
Mehul Damani, Isha Puri, Idan Shenfeld, and Jacob Andreas.
\newblock Right in the right way: Lm training with verifiable rewards and human demonstrations.
\newblock \emph{arXiv preprint arXiv:2607.01181}, 2026.

\bibitem[Fanconi et~al.(2025)Fanconi, Astorga, and van~der Schaar]{fanconi2025learning}
Claudio Fanconi, Nicol{\'a}s Astorga, and Mihaela van~der Schaar.
\newblock Learning reasoning rewards from expert demonstrations with inverse reinforcement learning.
\newblock \emph{arXiv preprint arXiv:2510.01857}, 2025.

\bibitem[Fu et~al.(2017)Fu, Luo, and Levine]{fu2017learning}
Justin Fu, Katie Luo, and Sergey Levine.
\newblock Learning robust rewards with adversarial inverse reinforcement learning.
\newblock \emph{arXiv preprint arXiv:1710.11248}, 2017.

\bibitem[Gao et~al.(2023)Gao, Schulman, and Hilton]{gao2023scaling}
Leo Gao, John Schulman, and Jacob Hilton.
\newblock Scaling laws for reward model overoptimization.
\newblock In \emph{International Conference on Machine Learning}, pp.\  10835--10866. PMLR, 2023.

\bibitem[{Google}(2026{\natexlab{a}})]{google2026gemini31pro}
{Google}.
\newblock Gemini 3.1 pro preview, 2026{\natexlab{a}}.
\newblock URL \url{https://ai.google.dev/gemini-api/docs/models/gemini-3.1-pro-preview}.

\bibitem[{Google}(2026{\natexlab{b}})]{google2026gemini37flash}
{Google}.
\newblock Gemini 3.7 flash, August 2026{\natexlab{b}}.
\newblock URL \url{https://ai.google.dev/gemini-api/docs/models/gemini-3.7-flash}.

\bibitem[Guo et~al.(2025)Guo, Yang, Zhang, Song, Wang, Zhu, Xu, Zhang, Ma, Bi, et~al.]{guo2025deepseek}
Daya Guo, Dejian Yang, Haowei Zhang, Junxiao Song, Peiyi Wang, Qihao Zhu, Runxin Xu, Ruoyu Zhang, Shirong Ma, Xiao Bi, et~al.
\newblock Deepseek-r1: Incentivizing reasoning capability in llms via reinforcement learning.
\newblock \emph{arXiv preprint arXiv:2501.12948}, 2025.

\bibitem[Ho \& Ermon(2016)Ho and Ermon]{ho2016generative}
Jonathan Ho and Stefano Ermon.
\newblock Generative adversarial imitation learning.
\newblock \emph{Advances in neural information processing systems}, 29, 2016.

\bibitem[Intellect(2025)]{primeintellect2025prime-rl}
Prime Intellect.
\newblock prime-rl, 2025.
\newblock URL \url{https://github.com/PrimeIntellect-ai/prime-rl}.

\bibitem[Jin et~al.(2025)Jin, Zhou, Zhang, Peng, Zhang, Dong, Pavone, Han, Hong, Che, et~al.]{jin2025your}
Can Jin, Yang Zhou, Qixin Zhang, Hongwu Peng, Di~Zhang, Zihan Dong, Marco Pavone, Ligong Han, Zhang-Wei Hong, Tong Che, et~al.
\newblock Your reward function for rl is your best prm for search: Unifying rl and search-based tts.
\newblock \emph{arXiv preprint arXiv:2508.14313}, 2025.

\bibitem[Li et~al.(2019)Li, Kiseleva, and De~Rijke]{li2019dialogue}
Ziming Li, Julia Kiseleva, and Maarten De~Rijke.
\newblock Dialogue generation: From imitation learning to inverse reinforcement learning.
\newblock In \emph{Proceedings of the AAAI Conference on Artificial Intelligence}, volume~33, pp.\  6722--6729, 2019.

\bibitem[Lightman et~al.(2024)Lightman, Kosaraju, Burda, Edwards, Baker, Lee, Leike, Schulman, Sutskever, and Cobbe]{lightman2024let}
Hunter Lightman, Vineet Kosaraju, Yuri Burda, Harrison Edwards, Bowen Baker, Teddy Lee, Jan Leike, John Schulman, Ilya Sutskever, and Karl Cobbe.
\newblock Let's verify step by step.
\newblock In \emph{International Conference on Learning Representations}, volume 2024, pp.\  39578--39601, 2024.

\bibitem[Lillicrap et~al.(2015)Lillicrap, Hunt, Pritzel, Heess, Erez, Tassa, Silver, and Wierstra]{lillicrap2015continuous}
Timothy~P Lillicrap, Jonathan~J Hunt, Alexander Pritzel, Nicolas Heess, Tom Erez, Yuval Tassa, David Silver, and Daan Wierstra.
\newblock Continuous control with deep reinforcement learning.
\newblock \emph{arXiv preprint arXiv:1509.02971}, 2015.

\bibitem[Ng et~al.(2000)Ng, Russell, et~al.]{ng2000algorithms}
Andrew~Y Ng, Stuart Russell, et~al.
\newblock Algorithms for inverse reinforcement learning.
\newblock In \emph{Icml}, volume~1, pp.\ ~2, 2000.

\bibitem[{OpenAI}(2024)]{openai2024embeddings}
{OpenAI}.
\newblock New embedding models and {API} updates, January 2024.
\newblock URL \url{https://openai.com/index/new-embedding-models-and-api-updates/}.

\bibitem[{OpenAI}(2026{\natexlab{a}})]{openai2026gpt56}
{OpenAI}.
\newblock {GPT-5.6}: Frontier intelligence that scales with your ambition, July 2026{\natexlab{a}}.
\newblock URL \url{https://openai.com/index/gpt-5-6/}.

\bibitem[{OpenAI}(2026{\natexlab{b}})]{openai2026gpt5systemcard}
{OpenAI}.
\newblock Gpt-5 system card.
\newblock \url{https://openai.com/index/gpt-5-system-card/}, 2026{\natexlab{b}}.
\newblock Accessed: 2026-08-25.

\bibitem[Ouyang et~al.(2022)Ouyang, Wu, Jiang, Almeida, Wainwright, Mishkin, Zhang, Agarwal, Slama, Ray, et~al.]{ouyang2022training}
Long Ouyang, Jeffrey Wu, Xu~Jiang, Diogo Almeida, Carroll Wainwright, Pamela Mishkin, Chong Zhang, Sandhini Agarwal, Katarina Slama, Alex Ray, et~al.
\newblock Training language models to follow instructions with human feedback.
\newblock \emph{Advances in neural information processing systems}, 35:\penalty0 27730--27744, 2022.

\bibitem[Pan(2025)]{pan2025countdown3to4}
Jiayi Pan.
\newblock Countdown-tasks-3to4.
\newblock \url{https://huggingface.co/datasets/Jiayi-Pan/Countdown-Tasks-3to4}, 2025.
\newblock Hugging Face dataset. Accessed: 2026-08-25.

\bibitem[Shao et~al.(2024)Shao, Wang, Zhu, Xu, Song, Bi, Zhang, Zhang, Li, Wu, et~al.]{shao2024deepseekmath}
Zhihong Shao, Peiyi Wang, Qihao Zhu, Runxin Xu, Junxiao Song, Xiao Bi, Haowei Zhang, Mingchuan Zhang, YK~Li, Yang Wu, et~al.
\newblock Deepseekmath: Pushing the limits of mathematical reasoning in open language models.
\newblock \emph{arXiv preprint arXiv:2402.03300}, 2024.

\bibitem[Shi et~al.(2018)Shi, Chen, Qiu, and Huang]{shi2018toward}
Zhan Shi, Xinchi Chen, Xipeng Qiu, and Xuanjing Huang.
\newblock Toward diverse text generation with inverse reinforcement learning.
\newblock \emph{arXiv preprint arXiv:1804.11258}, 2018.

\bibitem[Stiennon et~al.(2020)Stiennon, Ouyang, Wu, Ziegler, Lowe, Voss, Radford, Amodei, and Christiano]{stiennon2020learning}
Nisan Stiennon, Long Ouyang, Jeffrey Wu, Daniel Ziegler, Ryan Lowe, Chelsea Voss, Alec Radford, Dario Amodei, and Paul~F Christiano.
\newblock Learning to summarize with human feedback.
\newblock \emph{Advances in neural information processing systems}, 33:\penalty0 3008--3021, 2020.

\bibitem[Uesato et~al.(2022)Uesato, Kushman, Kumar, Song, Siegel, Wang, Creswell, Irving, and Higgins]{uesato2022solving}
Jonathan Uesato, Nate Kushman, Ramana Kumar, Francis Song, Noah Siegel, Lisa Wang, Antonia Creswell, Geoffrey Irving, and Irina Higgins.
\newblock Solving math word problems with process-and outcome-based feedback.
\newblock \emph{arXiv preprint arXiv:2211.14275}, 2022.

\bibitem[Wang et~al.(2024)Wang, Li, Shao, Xu, Dai, Li, Chen, Wu, and Sui]{wang2024math}
Peiyi Wang, Lei Li, Zhihong Shao, Runxin Xu, Damai Dai, Yifei Li, Deli Chen, Yu~Wu, and Zhifang Sui.
\newblock Math-shepherd: Verify and reinforce llms step-by-step without human annotations.
\newblock In \emph{Proceedings of the 62nd Annual Meeting of the Association for Computational Linguistics (Volume 1: Long Papers)}, pp.\  9426--9439, 2024.

\bibitem[Weininger(1988)]{weininger1988smiles}
David Weininger.
\newblock Smiles, a chemical language and information system. 1. introduction to methodology and encoding rules.
\newblock \emph{Journal of chemical information and computer sciences}, 28\penalty0 (1):\penalty0 31--36, 1988.

\bibitem[Wu et~al.(2021)Wu, Li, and Yu]{wu2021textgail}
Qingyang Wu, Lei Li, and Zhou Yu.
\newblock Textgail: Generative adversarial imitation learning for text generation.
\newblock In \emph{Proceedings of the AAAI Conference on Artificial Intelligence}, volume~35, pp.\  14067--14075, 2021.

\bibitem[Yang et~al.(2025)Yang, Li, Yang, Zhang, Hui, Zheng, Yu, Gao, Huang, Lv, et~al.]{yang2025qwen3}
An~Yang, Anfeng Li, Baosong Yang, Beichen Zhang, Binyuan Hui, Bo~Zheng, Bowen Yu, Chang Gao, Chengen Huang, Chenxu Lv, et~al.
\newblock Qwen3 technical report.
\newblock \emph{arXiv preprint arXiv:2505.09388}, 2025.

\bibitem[Ye et~al.(2025)Ye, Dong, Chi, Wu, Huang, and Wei]{ye2025black}
Tianzhu Ye, Li~Dong, Zewen Chi, Xun Wu, Shaohan Huang, and Furu Wei.
\newblock Black-box on-policy distillation of large language models.
\newblock \emph{arXiv preprint arXiv:2511.10643}, 2025.

\bibitem[Yuan et~al.(2024)Yuan, Li, Chen, Cui, Ding, Zhang, Zhou, Liu, and Peng]{yuan1981free}
Lifan Yuan, Wendi Li, Huayu Chen, Ganqu Cui, Ning Ding, Kaiyan Zhang, Bowen Zhou, Zhiyuan Liu, and Hao Peng.
\newblock Free process rewards without process labels, 2024.
\newblock \emph{URL https://arxiv. org/abs/2412.01981}, 2024.

\bibitem[Zha et~al.(2026)Zha, Gao, Shen, Hong, Boning, and Katabi]{zha2026rl}
Kaiwen Zha, Zhengqi Gao, Maohao Shen, Zhang-Wei Hong, Duane Boning, and Dina Katabi.
\newblock Rl tango: Reinforcing generator and verifier together for language reasoning.
\newblock \emph{Advances in Neural Information Processing Systems}, 38:\penalty0 119283--119313, 2026.

\bibitem[Zheng et~al.(2023)Zheng, Chiang, Sheng, Zhuang, Wu, Zhuang, Lin, Li, Li, Xing, et~al.]{zheng2023judging}
Lianmin Zheng, Wei-Lin Chiang, Ying Sheng, Siyuan Zhuang, Zhanghao Wu, Yonghao Zhuang, Zi~Lin, Zhuohan Li, Dacheng Li, Eric Xing, et~al.
\newblock Judging llm-as-a-judge with mt-bench and chatbot arena.
\newblock \emph{Advances in neural information processing systems}, 36:\penalty0 46595--46623, 2023.

\end{thebibliography}
